\documentclass[10pt,journal,compsoc]{IEEEtran}

\usepackage{times}
\usepackage{epsfig}
\usepackage{graphicx}
\usepackage{amsmath}
\usepackage{amssymb}
\usepackage{multirow}
\usepackage{cite}
\usepackage{caption}

\usepackage[bookmarks=false,colorlinks=true,linkcolor=black,citecolor=black,filecolor=black,urlcolor=black]{hyperref}

\newcommand{\etal}{\mbox{\emph{et al.\ }}}
\newcommand{\ie}{\mbox{\emph{i.e.}}}
\newcommand{\eg}{\mbox{\emph{e.g.}}}
\newlength\savewidth\newcommand\shline{\noalign{\global\savewidth\arrayrulewidth
  \global\arrayrulewidth 1pt}\hline\noalign{\global\arrayrulewidth\savewidth}}

\begin{document}

\title{Learning Image-Text Embeddings\\ with Instance Loss}

\author{Zhedong Zheng, 
        Liang Zheng,
		Michael Garrett,
        Yi Yang,
        Mingliang Xu,
		Yi-Dong Shen
\IEEEcompsocitemizethanks{
\IEEEcompsocthanksitem Zhedong Zheng, Liang Zheng and Yi Yang are with Centre for Artificial Intelligence, University of Technology Sydney, Australia. E-mail: \href{mailto:Zhedong.Zheng@student.uts.edu.au}{Zhedong.Zheng@student.uts.edu.au},
\href{mailto:Liang.Zheng@uts.edu.au}{Liang.Zheng@uts.edu.au},
\href{mailto:Yi.Yang@uts.edu.au}{Yi.Yang@uts.edu.au}
\IEEEcompsocthanksitem Michael Garrett is with CingleVue International Australia and Edith Cowan University, Australia. E-mail:
\href{mailto:michael.garrett@cinglevue.com}{michael.garrett@cinglevue.com}
\IEEEcompsocthanksitem Yi-Dong Shen is with Institute of Software, Chinese Academy of Sciences. Email:
\href{mailto:ydshen@ios.ac.cn}{ydshen@ios.ac.cn}
}
}

\markboth{Journal of \LaTeX\ Class Files,~Vol.~14, No.~8, August~2015}%
{Shell \MakeLowercase{\textit{et al.}}: Bare Demo of IEEEtran.cls for Computer Society Journals}

\IEEEcompsoctitleabstractindextext{%
\begin{abstract}
Matching images and sentences demands a fine understanding of both modalities. In this paper, we propose a new system to discriminatively embed the image and text to a shared visual-textual space. In this field, most existing works apply the ranking loss to pull the positive image / text pairs close and push the negative pairs apart from each other. 
However, directly deploying the ranking loss is hard for network learning, since it starts from the two heterogeneous features to build inter-modal relationship.
To address this problem, we propose the instance loss which explicitly considers the intra-modal data distribution. It is based on an unsupervised assumption that each image / text group can be viewed as a class. So the network can learn the fine granularity from every image/text group. The experiment shows that the instance loss offers better weight initialization for the ranking loss, so that more discriminative embeddings can be learned. Besides, existing works usually apply the off-the-shelf features, \ie, word2vec and fixed visual feature.
So in a minor contribution, this paper constructs an end-to-end dual-path convolutional network to learn the image and text representations. End-to-end learning allows the system to directly learn from the data and fully utilize the supervision. 
On two generic retrieval datasets (Flickr30k and MSCOCO),  experiments demonstrate that our method yields competitive accuracy compared to state-of-the-art methods. Moreover, in language based person retrieval, we improve the state of the art by a large margin. The code has been made publicly available.
\end{abstract}
}

\maketitle

\IEEEdisplaynotcompsoctitleabstractindextext

\IEEEpeerreviewmaketitle

\IEEEraisesectionheading{\section{Introduction}\label{sec:introduction}}
\IEEEPARstart{I}MAGE and text both contain very rich semantics but reside in heterogeneous modalities. Comparing to information retrieval within the same modality, matching image-text poses extra critical challenges, \emph{i.e.,} mapping images and text onto one shared feature space. For example, a model needs to distinguish between the ``black dog'', ``gray dog'' and ``two dogs'' in the text, and understand the visual differences in images depicting ``black dog'', ``gray dog'' and ``two dogs''. In this paper, given an unseen image (text) query, we aim to measure its semantic similarity with the text (image) instances in the database and retrieve the true matched texts (images) to the query. Considering the testing procedure, this task requires connecting the two modalities with robust representations. In the early times, some relatively small datasets were used, \eg, Wikipedia \cite{rasiwasia2010new} and Pascal Sentence \cite{rashtchian2010collecting}, which contain around 3,000 and 5,000 image-text pairs, respectively. In recent years, several large-scale datasets with more than 30,000 images, including MSCOCO \cite{lin2014microsoft} and Flickr30k \cite{young2014image}, have been introduced. Each image in these datasets is annotated with around five sentences.  These large datasets allow deep architectures to learn robust representations and provide challenging evaluation scenarios. 



During the past few years,  ranking loss is commonly used as the objective function \cite{frome2013devise,karpathy2014deep,ma2015multimodal,wang2016learning,nam2016dual,reed2016learning} for image-text representation learning. The ranking loss aims to make the distance between positive pairs smaller than that between negative pairs by a predefined margin. 
In image-text matching, every training pair contains a visual feature and a textual feature.
The ranking loss focuses on the distance between the two modalities. Its potential drawback is that it does not explicitly consider the feature distribution in a single modality. 
For example, when using ranking loss during training which does not distinguish between the slight differences in images, then given two testing images with slightly different semantics, the model may output similar descriptors for the two images. This is clearly undesirable for image / text matching considering the extremely fine granularity of this task. In our experiment, we observe that using the ranking loss alone in end-to-end training may cause the network to be stuck in a local minimum.

\begin{figure}[t]
\begin{center}
   \includegraphics[width=1\linewidth]{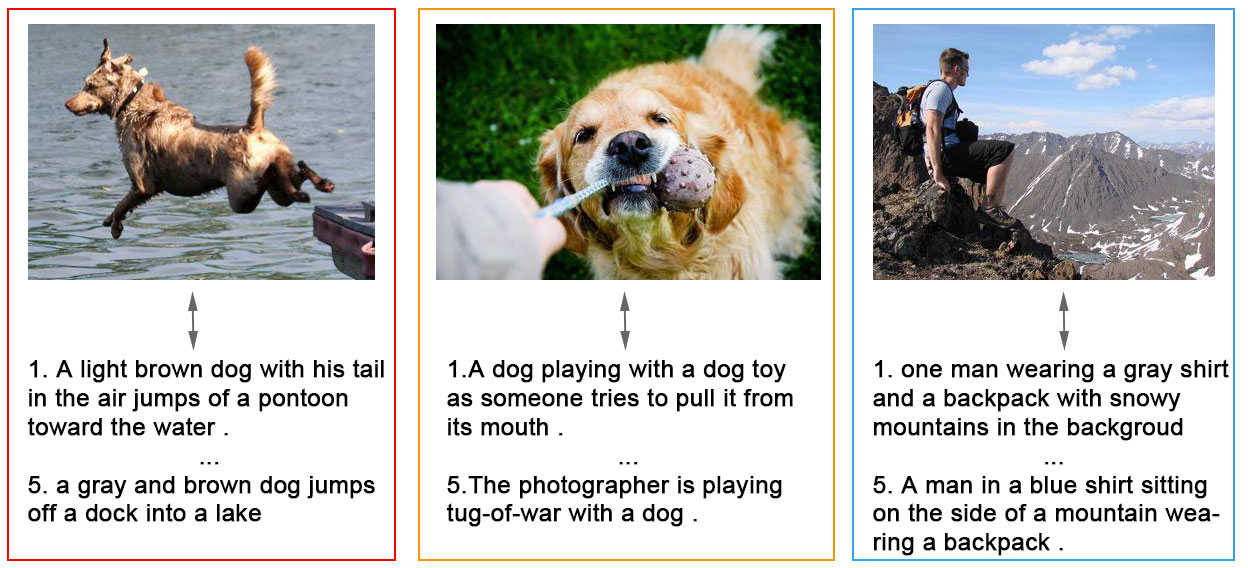}
\end{center}
   \caption{Motivation. We define an image / text group as an image with its associated sentences. We observe that an image / text group is more or less different from each other. Therefore, we view every image / text group as a distinct class during training, yielding the instance loss. }
\label{fig:0}
\end{figure}

What motivates us is the effectiveness of class labels in earlier years of cross-media retrieval \cite{sharma2012generalized,wang2013learning,wu2013cross,wei2017cross}. In these works, the class labels are annotated manually and during testing, the aim is to retrieve image / text belonging to the same class to the query. 
In light of this early practice, this paper explores the feasibility of ``class labels'' in image / text matching, which is an instance retrieval problem. Two differences exist between cross-media retrieval on the category level \cite{wu2013cross,wei2017cross} and on the instance level (considered in this paper). First, the true matches are those with the same category, and those with the exact same content with the query, respectively. That is to say, instance-level retrieval has a more strict matching criteria than category-level retrieval. Second, instance-level retrieval does not assume the existence of class labels. In this field of research, only image / text pairs are utilized during training. Given the intrinsic differences between the two tasks, it is non-trivial to directly transfer the experience from using class labels in category-level retrieval to instance-level retrieval. 

Without annotated class labels, how can we initiate the investigation of the underlying data structures in the image / text embedding space? In this paper, we name an image and its associated sentences an ``image / text group''. Our key assumption is that each ``image / text'' group is different from the others, and can be viewed as a distinct class (see Fig. \ref{fig:0}). 
So we propose a classification loss called instance loss to classify the image / text groups.
Using this unsupervised class labels as supervision, we aim to enforce the model to discriminate each two images and two sentences (from different groups). It helps to investigate the fine-grained difference in single modality (intra-modal) and provides a good initialization for ranking loss which is a driving force for end-to-end retrieval representation learning. In more details, using such an unsupervised assumption, we train the network to classify every image / text group with the softmax loss. In the experiment, we show that the instance loss which classifies a large number of classes, \ie, 113,287 image / text groups on MSCOCO \cite{lin2014microsoft}, is able to converge without any hyper-parameter tuning. Improved retrieval accuracy can  be observed as a result of instance loss.

In addition, we notice in the field of image-text matching that most recent works 
employ off-the-shelf deep models for image feature extraction \cite{mao2014deep,klein2015associating,lin2016leveraging,lev2016rnn,wang2016learning,nam2016dual,huang2016instance,niu2017hierarchical,wang2017learning,reed2016learning}. The fine-tuning strategy commonly seen in other computer vision tasks \cite{zhang2014part,chen2016deeplab,zheng2016person} is rarely adopted. 
A drawback of using off-the-shelf models is that these models are usually trained to classify objects into semantic categories \cite{krizhevsky2012imagenet,simonyan2014very,he2016deep}. The classification models are likely to miss image details such as color, number, and environment, which may convey critical visual cues for matching images and texts. For example, a model trained on ImageNet \cite{russakovsky2015imagenet} can correctly classify the three images as ``dog''; but it may not tell the difference between \emph{black dog} and \emph{gray dog}, or between \emph{one dog} and \emph{two dogs}. The ability to convey critical visual cues is a necessary component in instance-level image-text matching. Similar observations have been reported with regards to image captioning \cite{vinyals2017show}. Moreover, for the text feature, \emph{word2vec} \cite{mikolov2013efficient} is a popular choice in image-text matching \cite{klein2015associating,karpathy2015deep,wang2016learning,niu2017hierarchical}.
Aiming to model the context information, the \emph{word2vec} model is learned through a shallow network to predict neighboring words. However, the \emph{word2vec} model is initially trained on GoogleNews, which differs substantially from the text in the target dataset. As such, instead of using the off-the-shelf model, we explore the possibility of fine-tuning the \emph{word2vec} model using image-text matching datasets. 

\begin{figure*}[t]
\begin{center}
   \includegraphics[width=1\linewidth]{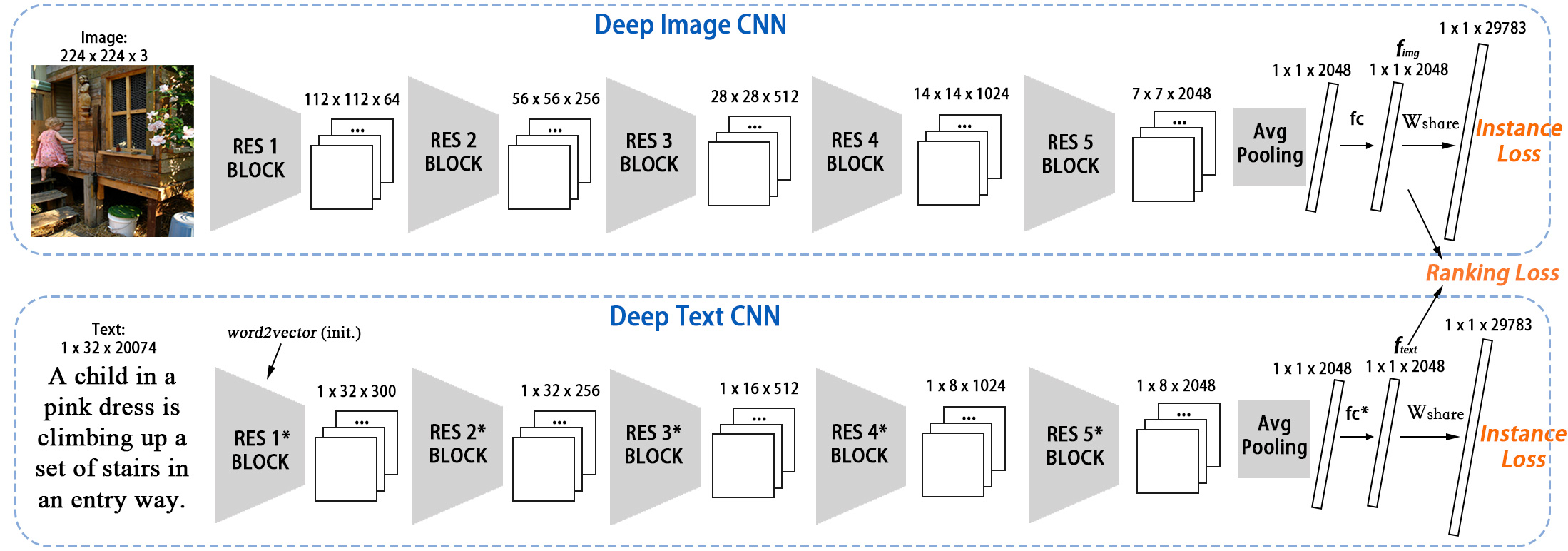}
\end{center}
   \caption{We learn the image and text representations by two convolutional neural networks, \ie, deep image CNN (top) and deep text CNN (bottom). The deep image CNN is a ResNet-50 model \cite{he2016deep} pre-trained on ImageNet. The deep text CNN is similar to the image CNN but with different basic blocks (see Fig. \ref{fig:3}). After the average pooling, we add one fully connected layer (input dim: $2,048$, output dim: $2,048$), one batchnorm layer, relu and one fully connected layer (input dim: $2,048$, output dim: $2,048$) in both image CNN and text CNN (We denote as fc and fc$^{*}$ in the figure, and the weights are not shared). Then we add a shared-weight $W_{share}$ classification layer (input dim: $2,048$, output dim: $29,783$). The objectives are the ranking loss and the proposed instance loss. On Flickr30k, for example, the model needs to classify 29,783 classes using instance loss.}
\label{fig:network}
\end{figure*}

Briefly, inspired by the effectiveness of class labels in early-time cross-media retrieval, we propose a similar practice in image-text matching called ``instance loss''. Instance loss works by providing better weight initialization for the ranking loss, thus producing more discriminative and robust image / text descriptions.  
Next, we also note that the pretrained CNN models may not meet the fine-grained requirement in image / text matching. So we construct a dual path CNN to extract image and text features directly from data rather. The network is end-to-end trainable and yields superior results to using features extracted from off-the-shelf models as input.  
Our contributions are summarized as follows: 

\begin{itemize}
\item To provide better weight initialization and regularize the dual-path CNN model, we propose a large-number classification loss called instance loss. The robustness and effectiveness of instance loss are demonstrated by classifying each image / text group into one of the 113,287 classes on MSCOCO \cite{lin2014microsoft}.
\item We propose a dual-path CNN model for visual-textual embedding learning  (see Fig. \ref{fig:network}). In contrast to the commonly used RNN+CNN model using fixed CNN features, the proposed CNN+CNN structure conducts efficient and effective end-to-end fine-tuning.
\item We obtain competitive accuracy compared with the state-of-the-art image-text matching methods on three large-scale datasets \ie, Flickr30k \cite{young2014image}, MSCOCO \cite{lin2014microsoft} and CUHK-PEDES \cite{li2017person}.
\end{itemize}

We note that Ma \etal also apply the CNN structure for text feature learning \cite{ma2015multimodal}. The main difference between our method and \cite{ma2015multimodal} is two-fold. First, Ma \etal \cite{ma2015multimodal}  use the ranking loss alone. In our method, we show that the proposed instance loss can further improve the result of ranking loss. Second, in \cite{ma2015multimodal}, four text CNN models are used to capture different semantic levels \ie, word, short phrase, long phrase and sentence. In this paper, only one text CNN model is used and the word-level input is considered. Our model uses the residual block shown in Fig. \ref{fig:3}, which combines low level information \ie, word, as well as high level inference to produce the final feature. In experiment (Table \ref{table:Flickr30k} and Table \ref{table:MSCOCO}), we show that using on the same image CNN (VGG-19), our method (with one text CNN) is superior to \cite{ma2015multimodal} with text model ensembles by a large margin.

The rest of this paper is organized as follows. Section \ref{sec:related work} reviews and discusses the related works. Section \ref{sec:model} describes the proposed Image-Text CNN Structure in detail, followed by the objective function in Section \ref{sec:loss}. Training policy is described in Section \ref{sec:two-stage}. Experimental results and comparisons are discussed in Section \ref{sec:experiments} and conclusions are in Section \ref{sec:conclusion}. Furthermore, some qualitative results are included in Appendix.


\section{Related Works} \label{sec:related work}
The image-text bidirectional retrieval requires both understanding images and sentences in detail. In this section, we discuss some related works. 

\textbf{Deep models for image recognition.}
Deep models have achieved success in computer vision. The convolutional neural network (CNN) won the ILSVRC12 competition \cite{russakovsky2015imagenet} by a large margin \cite{krizhevsky2012imagenet}. Later, VGGNet \cite{simonyan2014very} and ResNet \cite{he2016deep} further deepened the CNN and provide more insights into the network structure. 
In the field of image-text matching, most recent methods directly use fixed CNN features \cite{mao2014deep,klein2015associating,lin2016leveraging,lev2016rnn,wang2016learning,nam2016dual,huang2016instance,niu2017hierarchical,wang2017learning,reed2016learning} as input which are extracted from the models pre-trained on ImageNet. 
While it is efficient to fix the CNN features and learn a visual-textual common space, it may lose the fine-grained differences between the images. This motivates us to fine-tune the image CNN branch in the image-text matching to provide for more discriminative embedding learning.  

\textbf{Deep models for natural language understanding.} For natural language representation, \emph{word2vec}  \cite{mikolov2013efficient} is commonly used \cite{klein2015associating,karpathy2015deep,wang2016learning,niu2017hierarchical}. This model contains two hidden layers, which learns from the context information. In the application of image-text matching, Klein \etal \cite{klein2015associating} and Wang \etal \cite{wang2016learning} pool word vectors extracted from the fixed \emph{word2vec} model to form a sentence descriptor using Fisher vector encoding. Karpathy \etal \cite{karpathy2015deep} also utilize fixed word vectors as word-level input. With respect to this routine, this paper proposes an equivalent scheme to fine-tuning the \emph{word2vec} model, allowing the learned text representations to be adaptable to a specific task, which is, in our case, image-text matching.

Recurrent Neural Networks (RNN) are another common choice in natural language processing \cite{mikolov2010recurrent,wu2016google}. Mao \etal \cite{mao2014deep} employ a RNN to generate image captions. Similarly, Nam \etal \cite{nam2016dual} utilize directional LSTM \cite{hochreiter1997long} for text encoding, yielding state-of-the-art multi-modal retrieval accuracy. Conversely, our approach is inspired by recent CNN breakthroughs on natural language understanding. For example, Gehring \etal apply CNNs to conduct machine translation, yielding competitive results and more than 9.3x speedup on the GPU \cite{gehring2017convolutional}. There are also researchers who apply layer-by-layer CNNs for efficient text analysis \cite{hu2014convolutional,kim2014convolutional,zhang2015character,chen2015event}, obtaining competitive results in title recognition, event detection and text content matching. In this paper, in place of RNNs which are more commonly seen in image-text matching, we explore the usage of CNNs for text representation learning. 

\textbf{Multi-modal learning.}
There is a growing body of works on the interaction between multiple modalities. 
One line of works focus on \textbf{category-level retrieval} and leverage the category labels in the training set.
Sharma \etal \cite{sharma2012generalized} extend the Canonical Correlation Analysis \cite{hardoon2004canonical} (CCA) to learning class labels, and Wang \etal \cite{wang2013learning} learn the shared image-text space based on coupled input with class regression. Wu \etal \cite{wu2013cross} propose a bi-directional learning to rank for representation learning. In \cite{wei2017cross}, Wei \etal perform CNN fine-tuning by classifying categories on the training set and report an improved performance on image-text retrieval. Castrejon \etal deploy the multiple labels to learn the shared semantic space \cite{castrejon2016learning}.
The second line of works consider \textbf{instance-level retrieval} and, except for matched image-text pairs, do not provide any category label. Given a query, the retrieval objective is a specific image or related sentences. Some works apply the auto-encoder to project high-dimensional features from different modalities onto a common low-dimensional latent space \cite{wang2014effective,eisenschtat2016linking}. In \cite{zhang2017discriminative}, Zhang \etal consider the verification loss, using a binary classifier to classify the true matches and false matches. 
Another line of works widely apply the ranking loss for instance-level retrieval \cite{frome2013devise,karpathy2014deep,ma2015multimodal,wang2016learning,nam2016dual,reed2016learning}. Karpathy \etal propose a part-to-part matching approach using a global ranking objective \cite{karpathy2014deep}. 
The ``SPE'' proposed in \cite{wang2016learning} extends the ranking loss with structure-preserving constraints. SPE is similar to our work in that both works consider the intra-modal distance. Nevertheless, our work differs significantly from SPE. SPE enforces the model to rank the texts, \ie, considering the feature separability within the text modality only.
In comparison, with the proposed instance loss, our method jointly discriminates the two modalities, \ie, images and their associated texts.


Briefly, we focus on instance-level retrieval and propose the instance loss, a novel contribution to the cross-modality community. It views each training image / text group as a distinct class and uses the softmax loss for model training. The assumption is unsupervised. We show that this method converges well and yields consistent improvement. 



\section{Proposed CNN Structure} \label{sec:model}
In this paper, we propose a dual path CNN to simultaneously learn visual and textual representations in an end-to-end fashion, consisting of a deep image CNN for image input and one deep text CNN for sentence input. The entire network only contains four components, \ie, convolution, pooling, ReLU and batch normalisation. Compared to many previous methods which use off-the-shelf image CNNs \cite{mao2014deep,klein2015associating,lin2016leveraging,lev2016rnn,wang2016learning,nam2016dual,huang2016instance,niu2017hierarchical,wang2017learning,reed2016learning}, end-to-end fine-tuning is superior in learning representations that encode image details (see Fig. \ref{fig:network}).

\subsection{Deep Image CNN} 
We use ResNet-50 \cite{he2016deep} pre-trained on ImageNet \cite{krizhevsky2012imagenet} as a basic model (the final 1000-classification layer is removed) before conducting fine-tuning for visual feature learning. Given an input image of size $224 \times 224$, a forward pass of the network produces a $2,048$-dimension feature vector. Followed by this feature, we add one fully-connected layer (input dim: $2,048$, output dim: $2,048$), one batch normalization, relu and one fully-connected layer (input dim: $2,048$, output dim: $2,048$). We denote the final $2,048$-dim vector $f_{img}$ as the visual descriptor of the input $I$. The forward pass process of the CNN, which is a non-linear function, is represented by function $\mathcal{F}_{img}(\cdot)$ defined as:
\begin{equation}
\label{eq:feature_img}
f_{img} = \mathcal{F}_{img}(I).
\end{equation}

\subsection{Deep Text CNN}\label{sec:text_cnn}
\textbf{Text processing.} Next, we describe our text processing method and the text CNN structure.
Given a sentence, we first convert it into code $T$ of size $n \times d$, where $n$ is the length of the sentence, and $d$ denotes the size of the dictionary. $T$ is used as the input for the text CNN.  We use $word2vec$ \cite{mikolov2013efficient} as a general dictionary to filter out rare words; if a word does not appear in the $word2vec$ dictionary (3,000,000 words), it is discarded. For Flickr30k, we eventually use $d = 20,074$ words as the dictionary. Every word in Flickr30k thus can find an index $l\in[1,d]$ in the dictionary; for instance, a sentence of 18 words can be converted to $18 \times d$ matrix. 
The text input $T$ can thus be formulated as: 
\begin{equation}
T(i,j) = 
   \begin{cases}
   1 &\mbox{if }j = l_i\\
   0 &\mbox{otherwise}
   \end{cases},
\end{equation}
where $i\in[1,18], j\in[1,d]$. The text CNN needs a fixed-length input. We set a fixed length $32$ in this paper because about 98\% sentences contain less than $32$ words. If the length of the sentence is shorter than $32$, we pad with zeros to the columns of $T$. If the length of the sentence is longer than $32$, we clip the final several words. Now we obtain the $32\times d$ sentence code $T$. We further reshape $T$ into the $1 \times 32 \times d$ format, which can be considered as height, width and channel known in the image CNNs \cite{he2016deep,krizhevsky2012imagenet}.

\begin{figure}[t]
\begin{center}
   \includegraphics[width=1\linewidth]{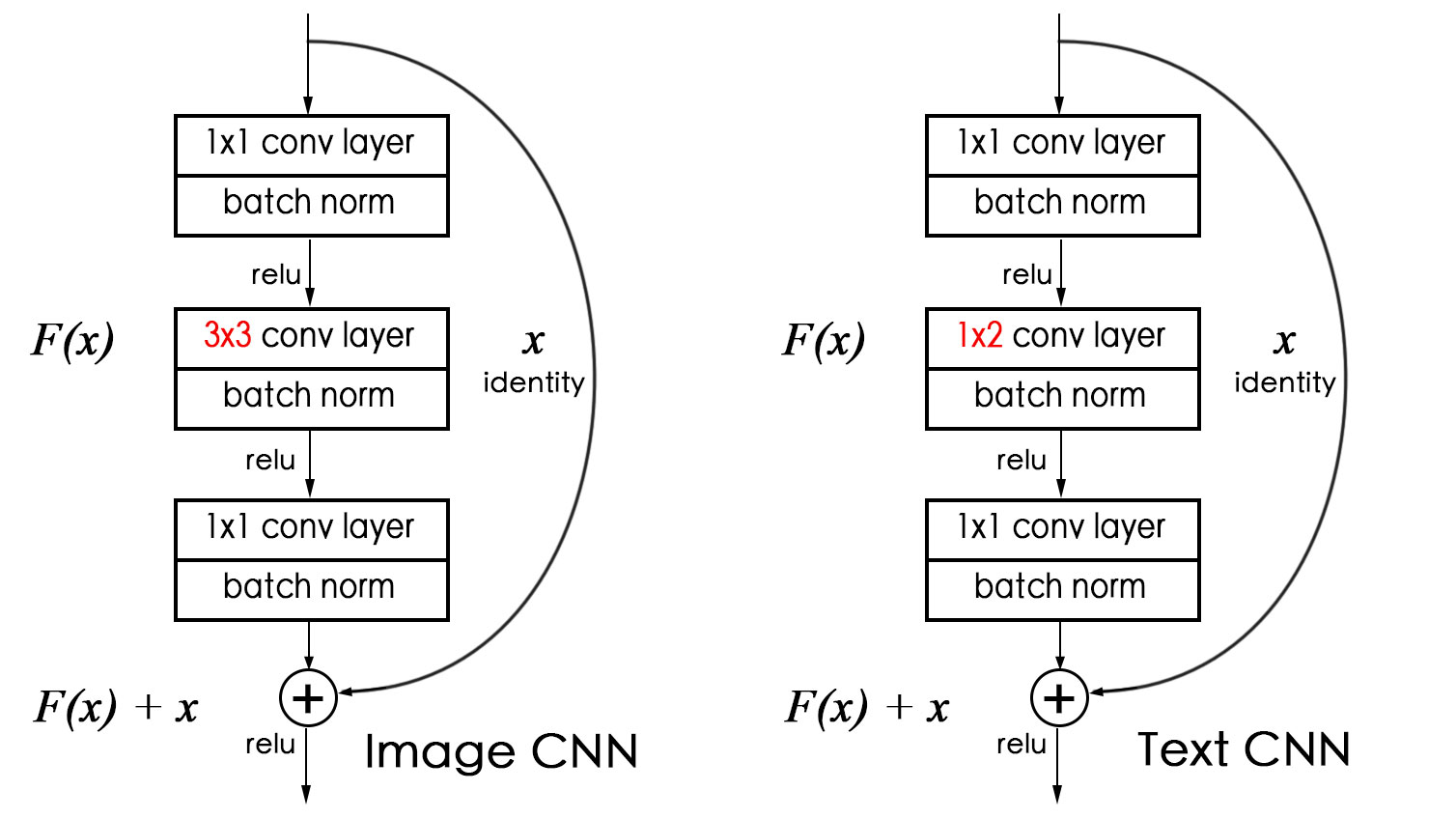}
\end{center}
   \caption{The basic block of deep image CNN and deep text CNN. Similar with the local pattern of the images, the neighbor words in the sentence may contains important clues. The filter size in the image CNN is $3\times3$ with height and width padding; the filter size in the text CNN is $1\times2$ with length padding. Besides, we also use a shortcut connection, which helps to train a deep convolutional network \cite{he2016deep}. The output $\mathcal{F}(x)+x$ has the same size with the input $x$.}
\label{fig:3}
\end{figure}

\emph{Position shift.}
We are motivated by the jittering operation in the image CNN training. For text CNN, we apply a data augmentation policy called position shift. 
In a baseline approach, if the sentence length $n$ is shorter than the standard input length $32$, a straightforward idea is to pad zeros at the end of the sentence, called \emph{left alignment}. In the proposed position shift approach, we pad a random number of zeros at the beginning and the end of a sentence. In this manner, shift variations are contained in the text representation, so that the learned embeddings are more robust. In the experiment, we observe that position shift is of importance to the performance. 

\textbf{Deep text CNN.} In the text CNN, filter size of the first convolution layer is $1 \times 1 \times d \times 300$, which can be viewed as a lookup table. Using the first convolutional layer, a sentence is converted to the word vector as follows. Given input $T$ of $1 \times 32 \times d$, the first convolution layer results in a tensor of size $1 \times 32 \times 300$. There are two methods to initialize the first convolutional layer: 1) random initialization \cite{glorot2010understanding}, and 2) using the $d\times 300$-dim matrix from \emph{word2vec} for initialization. In the experiment, we observe that \emph{word2vec} initialization is superior to the random initialization.

For the rest of the text CNN, similar residual blocks are used as per the image CNN (see Fig. \ref{fig:3}). Similar to the local pattern in the image CNN, every two neighbor components may form a phrase containing content information. We set the filter size of convolution layers in basic text block to $1 \times 2$. Additionally, we add the shortcut connection in the basic block, which has been demonstrated to help training deep neural networks  \cite{he2016deep}. We apply basic blocks with a short connection to form the deep textual network (see Fig. \ref{fig:network}). The number of blocks is consistent with the ResNet-50 model in the visual branch. Given a sentence matrix $T$, its text descriptor $f_{text}$ can be extract in an end-to-end manner from the text CNN $\mathcal{F}_{text}(\cdot)$:
\begin{equation}
\label{eq:feature_text}
f_{text} = \mathcal{F}_{text}(T).
\end{equation}

\section{Proposed Instance Loss} \label{sec:loss}
In this paper, two types of losses are used, \emph{i.e.,} the standard ranking loss and the proposed instance loss. In Section \ref{sec:ranking_loss}, we briefly review the formulation of the ranking loss and discuss the limitation of the ranking loss. Section \ref{sec:instance} describes the motivation and the formulation of the instance loss followed by a discussion. The differences between instance loss and ranking loss are discussed, and some primary experiments show the feasibility of instance loss. In Section \ref{sec:convergence}, training convergence of the instance loss is discussed.

\subsection{Ranking Loss Review} \label{sec:ranking_loss}
Ranking loss is a widely used objective function for retrieval problems. 
We use the cosine distance $D(f_{x_i},f_{x_j})=\frac{f_{x_i}}{||f_{x_i}||_2} \times \frac{f_{x_j}}{||f_{x_j}||_2}$ to measure the similarity between two samples, where $f$ is the feature of a sample, and $||\cdot||_2$ denotes the L2-norm. The distance value $D(f_{x_i},f_{x_j}) \in [-1,1]$.

To effectively account for two modalities, we follow the ranking loss formulation as in some previous works \cite{karpathy2014deep,nam2016dual}. Here, $I$ denotes the visual input, and $T$ denotes the text input. 
Given a quadric input $(I_a,T_a,I_n,T_n)$, where $I_a, T_a$ describe the same image / text group, $I_n, T_n$ are negative samples, ranking loss can be written as, 
\begin{multline}
\label{eq:rank_loss}
L_{rank} = \overbrace{max[0, \alpha-(D(f_{I_a},f_{T_a})-D(f_{I_a},f_{T_n}))]}^{image \ anchor}\\ 
 + \underbrace{max[0,\alpha-(D(f_{T_a},f_{I_a})-D(f_{T_a},f_{I_n}))]}_{text \ anchor},
\end{multline}
where $D(\cdot,\cdot)$ is the cosine similarity, and $\alpha$ is a margin.
Given an image query $I_a$, the similarity score of the correct text matching should be higher. Similarly, if we use sentence query $T_a$, we expect the correct image content should be ranked higher. Ranking loss explicitly builds the relationship between the image and text. 


\textbf{Limitations of ranking loss.} Although widely used, ranking loss has a potential drawback for the application of image-text matching. According to Eq. \ref{eq:rank_loss}, every pair contains a visual feature and a textual feature. The ranking loss focuses on the distance between the two modalities. So the potential drawback is that the ranking loss does not explicitly consider the feature distribution in a single modality. For instance, given two testing images with slightly different semantics, the model may output similar features. It is clearly undesirable for the extremely fine granularity of this task. 
In the experiment, using ranking loss alone is prone to get stuck in a local minimum (as to be shown in Fig. \ref{fig:loss-compare} and Table \ref{table:instance>Rank}).

\subsection{Instance Loss} \label{sec:instance}

\textbf{Motivation.}
Some early works use coarse-grain category \ie, art, biology, and sport, as the training supervision \cite{sharma2012generalized,wang2013learning,wei2017cross}. The multi-class classification loss has shown a good performance. But for instance-level retrieval, the classification loss has not been used. There may be two reasons. First, the category-level annotations are missing for most large-scale datasets. Second, if we use the category to train the model, it forces different instances, \ie, black dog, and white dogs, to the same class. It may compromise the CNN to learn the fine-grained difference. 

\begin{figure*}[t]
\begin{center}
   \includegraphics[width=1\linewidth]{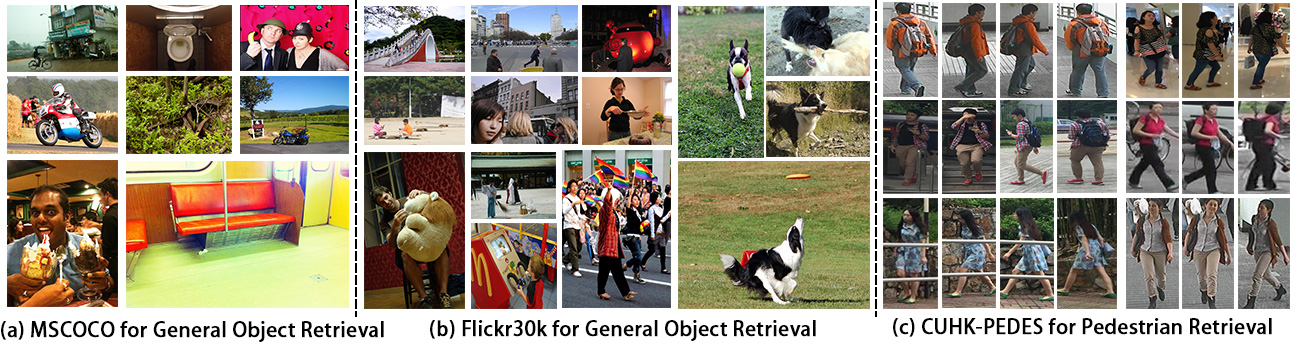}
\end{center}
   \caption{Sample images in the three datasets. For the MSCOCO and Flickr30k datasets, we view every image and its captions as an image / text group. For CUHK-PEDES, we view every identity (with several images and captions) as a class. }
\label{fig:6}
\end{figure*}

In this paper, we propose the instance loss for instance-level image-text matching. We define an image and its related text descriptions as an image / text group. In specific applications such as language-based person retrieval \cite{li2017identity,li2017person}, an image / text group is defined as images and their descriptions which depict the same person (see Fig. \ref{fig:6}). Based on image / text groups, our assumption is that each image / text group is distinct (duplicates have been removed in the datasets). Under such assumption, we view each image / text group as a class. So in essence, \emph{instance loss is a softmax loss which classifies an image / text group into one of a large number of classes.} We want the trained model can tell the difference between every two images as well as every two sentences (from different groups). Formally, we define instance loss below.

\textbf{Formulation.}
For two modalities, we formulate two classification objectives as follows,
\begin{align}
P_{visual} &= softmax(W_{share}^{T} f_{img}), \\ \label{eq:vW}
L_{visual} &= -\log(P_{visual}(c)), \\  
P_{textual} &= softmax(W_{share}^{T} f_{text}), \\ \label{eq:tW}
L_{textual} &= -\log(P_{text}(c)),
\end{align}
where $f_{img}$ and $f_{text}$ are image and text features defined in Eq. \ref{eq:feature_img} and Eq. \ref{eq:feature_text}, respectively. $W_{share}$ is the parameter of the final fully connected layer (Fig. \ref{fig:network}). It can be viewed as concatenated weights $W_{share} = [W_1, W_2, ..., W_{29783}].$ Every weight $W_i$ is a 2048-dim vector. $L$ denotes the loss and $P$ denotes the probability over all classes. $P(c)$ is the predicted possibility of the right class $c$. \textbf{Here we enforce shared weight $W_{share}$ in the final fully connected layer for the two modalities, because otherwise the learned image and text features may exist in totally different subspaces.}

As to be described in Section \ref{sec:two-stage}, in the first training stage, the ranking loss is not used. We only use the instance loss; in the second training stage, both losses are used.
The final loss function is a combination of the ranking loss and the instance loss, defined as,
\begin{equation}
\label{eq:final_loss}
L = \lambda_1 L_{rank} + \lambda_2 L_{visual} + \lambda_3 L_{textual},
\end{equation}
where $\lambda_1$, $\lambda_2$, $\lambda_3$ are predefined weights for different losses.

\begin{figure}[t]
 \begin{center}
    \includegraphics[width=1\linewidth]{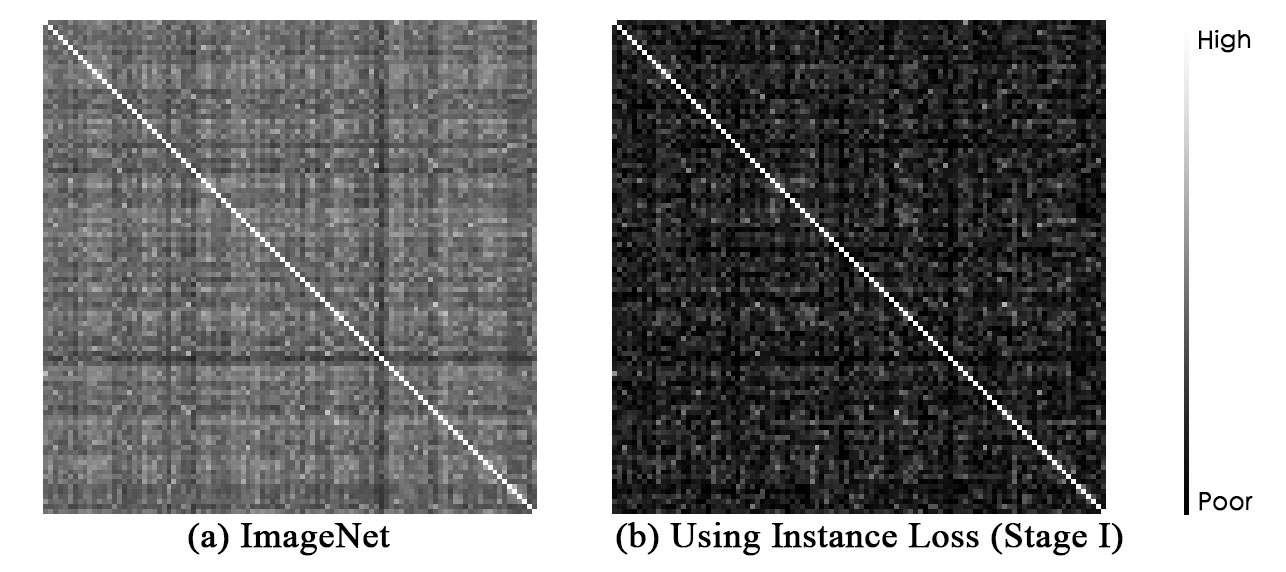}
 \end{center}
    \caption{We extract image features (2,048-dim) from a randomly selected 100 images in the Flickr30k validation set, using the ImageNet pre-trained ResNet-50 model and our model (after Stage I), respectively. We visualize the $100 \times 100$ Pearson's correlation. Lower Pearson's correlation between two features indicates higher orthogonality. The instance loss encourages the model to learn the difference between images.}
 \label{fig:imagenet}
\end{figure}

\label{discussion}
\textbf{Discussion.} First, we show that instance loss provides better weight initialization than the ImageNet pretrained model. To prove this, we compare the image features from the off-the-self model pre-trained on ImageNet and the model trained with instance loss. Since the proposed instance loss explicitly considers the intra-modal distance, we observe that the feature correlation between two images is smaller after training with the instance loss (see Fig. \ref{fig:imagenet}(b)). 
In fact, the instance loss encourages the model to find the fine-grained image details such as ball, stick, and frisbee to discriminate between image / text groups with similar semantics. We visualize the dog retrieval results in Fig. \ref{fig:dogs}. Our model can be well generalized to the test set and still sensitive to the subtle differences.

\begin{figure*}[t]
\begin{center}
   \includegraphics[width=1\linewidth]{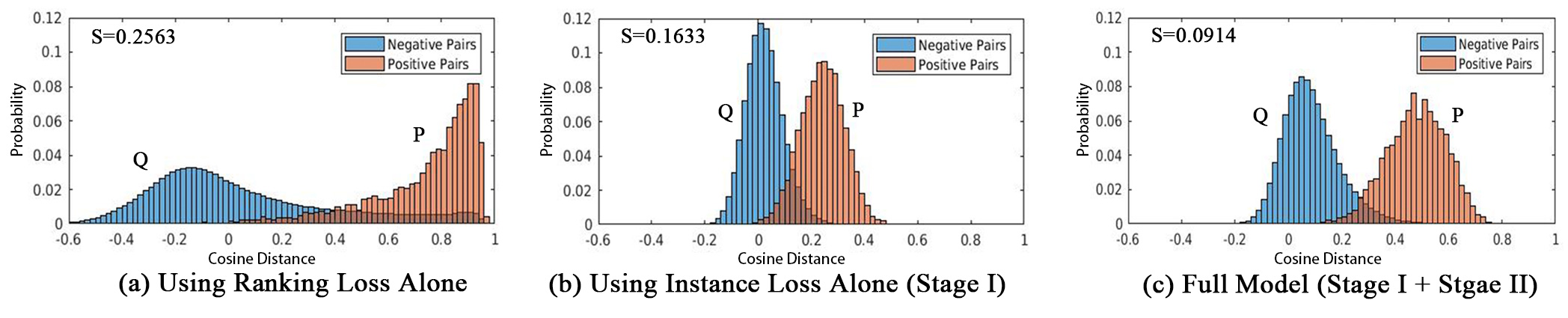}
\end{center}
   \caption{The similarity (cosine distance) distribution of the positive pairs $P$ and negative pairs $Q$ on Flickr30k validation dataset. We show the result obtained by (a) using ranking loss alone, (b) using instance loss alone and (c) full model (instance loss + ranking loss), respectively. Indicator $S$ is calculated as the overlapping area between $P$ and $Q$ (defined in Section \ref{sec:instance}, lower is better). Through comparing their $S$ values, the performance of the three methods is: ``Full Model'' $>$ ``Using Instance Loss Alone'' $>$ ``Using Ranking Loss Alone''.}
\label{fig:loss-compare}
\end{figure*}

\begin{figure}[t]
 \begin{center}
    \includegraphics[width=1\linewidth]{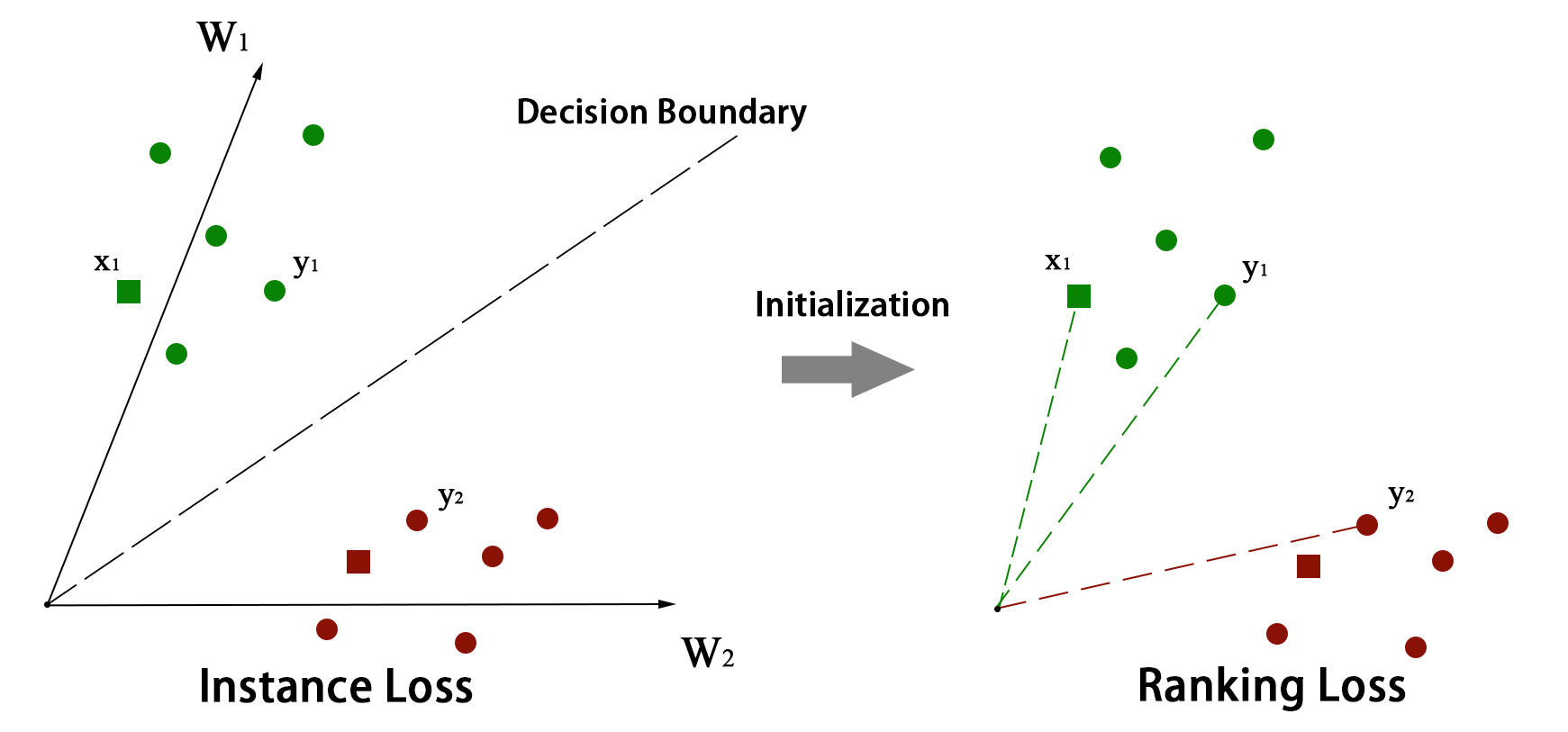}
 \end{center}
    \caption{Geometric Interpretation. The proposed instance loss leads to a decent weight initialization for ranking loss + instance loss in Stage II.}
 \label{fig:loss}
\end{figure}

Second, we provide an example of two classes to describe the working mechanism of instance loss (Fig. \ref{fig:loss}). $W_{share} = [W_1, W_2].$
Given image $x_1$ which belongs to the first class, the softmax loss function informs the constraint of $W_1^Tf_{x_1}> W_2^Tf_{x_1}$. Similarly, if $y_1$ is an input sentence belonging to the first class, the softmax loss will lead to the constraint of $W_1^Tf_{y_1}> W_2^Tf_{y_1}.$ The decision boundary indicates equal probability to be classified into the first class and the second class. Since the image and text embedding networks share the same final weight $W_{share}$, the features of the same image / text group will be close to each other in the embedding space; the data points from different image / text groups will be pushed away from each other. Therefore, after training with the instance loss, the data points will usually locate on the either side of the decision boundary.
In this manner, the image / text groups can be separated in the feature space despite of the fine-grained differences among them. This property,  as shown in the Fig. \ref{fig:loss} (\emph{right}), will provide better weight initialization for the subsequent training with both the ranking loss and instance loss. 

Third, we demonstrate that using the instance loss alone can lead to a decent initialization. To validate this point, we plot the distribution $P$ of the intra-modal intra-class similarity $D_{p}=D(f_{x_{i}}, f_{y_{i}})$ and the distribution $Q$ of the intra-modal inter-class similarity $D_{n}=D(f_{x_{i}},f_{y_{j}}) (j\neq i)$ on Flickr30k validation set (Fig. \ref{fig:loss-compare}(b)). 
We observe that, using instance loss alone, in most cases, leads to $D_p>D_n$ by a margin. The mean of $D_p$ equals to $0.2405$ while the mean of $D_n$ is $0.0237$.

Fourth, using the ranking loss alone achieves a relatively large margin between the positive pairs and negative pairs but there also exist many ``hard'' negative pairs (Fig. \ref{fig:loss-compare}(a)).  These ``hard'' negative pairs usually have a high similarity which compromises the matching performance of the true matches. Due to the potential drawback of the ranking loss mentioned in Section \ref{sec:ranking_loss}, the image / text with slightly difference may have the similar feature, which result in the ``hard'' negative samples. To quantitatively compare the three models, we propose a simple indicator function,
\begin{align}
S =  \int_{-1}^{1} min(P(x), Q(x))dx,
\end{align}
which encodes the overlapping area of $P$ and $Q$ over the range of cosine similarity $[-1,1]$. Indicator $S\in [0, 1]$. The smaller $S$ is, the better the positive pairs and negative pairs are separated, and thus the better retrieval performance. $S=1$ indicates the case where the two distributions, $P$ and $Q$ are completely overlapping. Under this \emph{worst case}, the positive pairs cannot be distinguished from the negative ones, and the retrieval performance is random. To the other extreme, $S=0$ indicates that the positive pairs and negative pairs are perfectly separable: all the similarity scores of the positive pairs are larger than the similarity scores of the negative pairs. In this \emph{best case}, the retrieval precision and recall are both 100\%.
Therefore, a lower indicator score $S$ indicates a better retrieval system. 

In our experiment (Fig. \ref{fig:loss-compare}), the indicator scores of the three models are $S_{rank} = 0.2563$, $S_{instance} = 0.1633$ and $S_{full} = 0.0914$, respectively. It clearly demonstrates that in terms of the extent of feature separability: ``Full Model'' $>$ ``Using Instance Loss Alone'' $>$ ``Using Ranking loss Alone''. With the indicator function, we quantitatively show that using ranking loss alone produces more hard negative pairs than the proposed two competing methods, which compromises the matching performance of the ranking loss. In comparison, using instance loss alone produces a smaller $S$ value, suggesting a better feature separability of the trained model. Importantly, when the two losses, \ie, ranking loss and instance loss, are combined, our full model has the smallest $S$ value, indicating the fewest hard negative samples and the best retrieval accuracy among the three methods.

For the retrieval performance, using the instance loss alone can lead to a competitive accuracy in the experiment (Table \ref{table:instance>Rank}). The effect of the instance loss is two-fold. In the first training stage, when used alone, it pre-trains the  text CNN and fine-tunes the two fully-connected layers (and one batchnorm layer) of image CNN so that ranking loss can arrive at a better optimization for both modalities in the second stage (Fig. \ref{fig:loss}). In the second training stage, when used together with ranking loss, it exhibits a regularization effect on the ranking loss. 

\begin{figure}[t]
\begin{center}
   \includegraphics[width=1\linewidth]{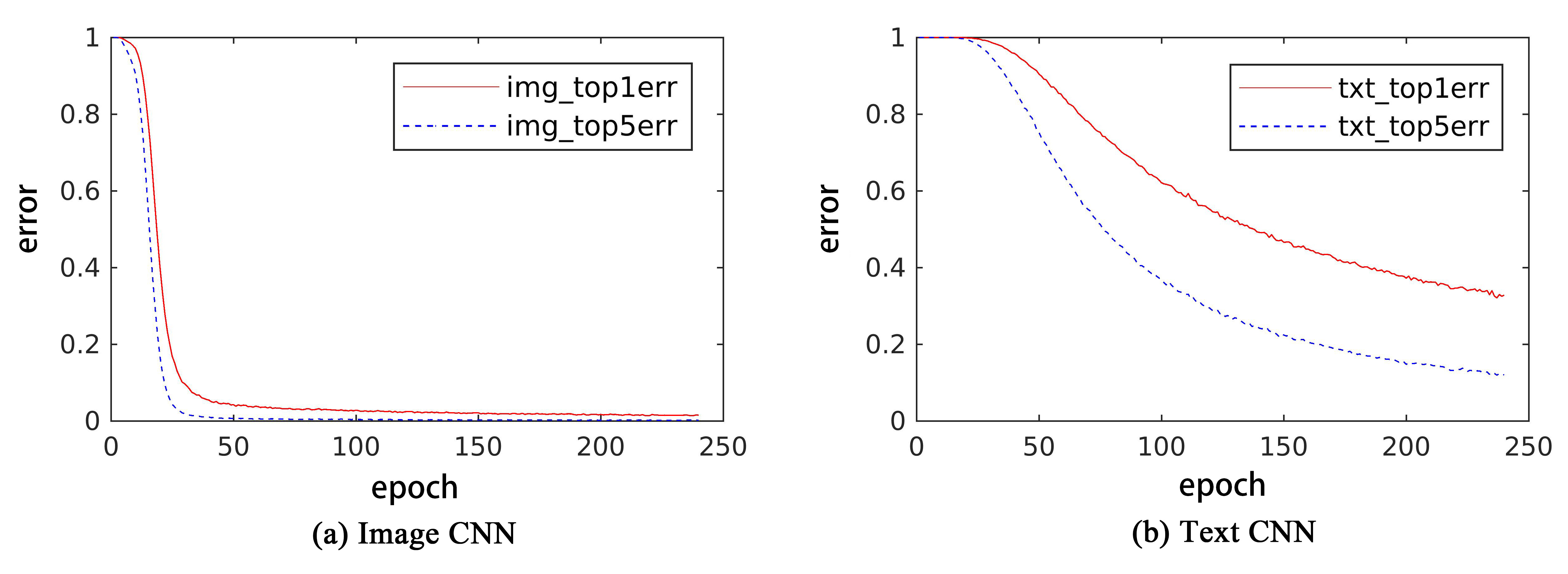}
\end{center}
   \caption{Classification error curves when training on Flickr30k. The image CNN (a) and text CNN (b) converge well with 29,783 training classes (image / text groups).}
\label{fig:converge}
\end{figure}

\subsection{Training Convergence of Instance Loss}\label{sec:convergence}
The instance loss views every image / text group as a class, so the number of training classes is usually large. For instance, we have 29,783 classes when training on Flickr30k. In Fig. \ref{fig:converge}, we show the training error curves of the image CNN and text CNN during training. 
We observe that the image CNN converges faster (Fig. \ref{fig:converge}(a)) because the image CNN is pretrained on ImageNet. Text CNN converges more slowly because most part of it is trained from scratch, but it still begins to learn something after 20 epochs, and finally converges after 240 epochs. 

On the other hand, the convergence property is evidenced by some previous works. To our knowledge, some practices also suffer from limited data per class, because manually annotating data is usually expensive. For example, in person re-ID, CUHK03 dataset \cite{li2014deepreid} has 9.6 training samples per class; VIPeR dataset \cite{gray2007evaluating} has 2 training samples per class. The previous works \cite{zheng2016discriminatively, qian2017multi} on CUHK03 and VIPeR show that the CNN classification model can be well trained as long as each class has more than a couple of training samples. In our case, there are usually 6 positive training samples per class (1 image and 5 sentences). In the experiment, despite of the limited training data, the learned model has a good generalization ability on the validation set and test set, which accords with existing experience \cite{zheng2016discriminatively, qian2017multi}. 
 


\section{A Two-Stage Training Procedure}\label{sec:two-stage}
We describe the training policy in this section. We split the training procedure into two stages. In the experiment, we show this policy helps the training.

\textbf{Stage I:} In this stage, we fix the pre-trained weights in the image CNN and use the proposed instance loss to tune the remaining part. The main reason is that most weights of the text CNN are learned from scratch. If we train the image and text CNNs simultaneously, the text CNN may compromise the pre-trained image CNN. We only use the proposed instance loss in this stage ($\lambda_1 = 0, \lambda_2 =1, \lambda_3 = 1$). It can provide a good initialization for the ranking loss. We note that even after Stage I, our network can achieve competitive results compared to previous works using off-the-shelf CNNs. 

\textbf{Stage II:} After Stage I converges, we start Stage II for end-to-end fine-tuning of the \emph{entire network}. Note that the weights of the image CNN are also fine-tuned. In this stage, we combine the instance loss with the ranking loss ($\lambda_1 = 1, \lambda_2 =1, \lambda_3 = 1$), so that both classification and ranking errors are considered. In Section \ref{stage2}, we study the mechanism of the two losses. It can be observed that in Stage II, instance loss and ranking loss are complementary, thus further improving the retrieval result. Instance loss still regularizes the model and provides more attentions to discriminate the images and sentences. After Stage II (end-to-end fine-tuning), another round of performance improvement can be observed, and we achieve even more competitive performance.


\setlength{\tabcolsep}{3.5pt}
\begin{table*}
\footnotesize
\begin{center}
\begin{tabular}{l|c|c|cccc|cccc}
\hline
\multirow{2}{*}{Method} & \multirow{2}{*}{Visual} & \multirow{2}{*}{Textual} & \multicolumn{4}{c|}{Image Query} & \multicolumn{4}{c}{Text Query}\\
   & & & R@1 & R@5 & R@10 & Med & R@1 & R@5 & R@10 & Med $r$\\
\shline
DeVise \cite{frome2013devise} & ft AlexNet & ft skip-gram &4.5 & 18.1 & 29.2 & 26 & 6.7 & 21.9 & 32.7 & 25 \\
Deep Fragment \cite{karpathy2014deep} & ft RCNN & fixed word vector from \cite{huang2012improving} &16.4 & 40.2 & 54.7 & 8 & 10.3 & 31.4 & 44.5 & 13 \\
DCCA \cite{yan2015deep} & ft AlexNet  & TF-IDF &16.7 & 39.3 & 52.9 & 8 & 12.6 & 31.0 & 43.0 & 15 \\
DVSA \cite{karpathy2015deep} &  ft RCNN (init. on Detection) & w2v + ft RNN &22.2 & 48.2 & 61.4 & 4.8 & 15.2 & 37.7 & 50.5 & 9.2 \\
LRCN \cite{donahue2015long} & ft VGG-16  & ft RNN & 23.6 & 46.6 & 58.3 & 7 & 17.5 & 40.3 & 50.8 & 9 \\
m-CNN \cite{ma2015multimodal} & ft VGG-19 & 4 $\times$ ft CNN &33.6 & 64.1 & 74.9 & 3 & 26.2 & 56.3 & 69.6 & 4\\
VQA-A \cite{lin2016leveraging} & fixed VGG-19 & ft RNN&33.9 & 62.5 & 74.5 & - & 24.9 & 52.6 & 64.8 & - \\
GMM-FV \cite{klein2015associating} &  fixed VGG-16 & w2v + GMM + HGLMM &35.0 & 62.0 & 73.8 & 3 & 25.0 & 52.7 & 66.0 & 5\\
m-RNN \cite{mao2014deep} & fixed VGG-16 & ft RNN &35.4 & 63.8 & 73.7 & 3 & 22.8 & 50.7 & 63.1 & 5 \\
RNN-FV \cite{lev2016rnn} &  fixed VGG-19 & feature from  \cite{klein2015associating} &35.6 & 62.5 & 74.2 & 3 & 27.4 & 55.9 & 70.0 & 4 \\
HM-LSTM \cite{niu2017hierarchical} & fixed RCNN from \cite{karpathy2015deep} & w2v + ft RNN &38.1 & - & 76.5 & 3 & 27.7 & - & 68.8 & 4 \\
SPE \cite{wang2016learning} &  fixed VGG-19 & w2v + HGLMM &40.3 & 68.9 & 79.9 & - & 29.7 & 60.1 & 72.1 & - \\
sm-LSTM \cite{huang2016instance} & fixed VGG-19 & ft RNN &42.5 & 71.9 & 81.5 & 2 & 30.2 & 60.4 & 72.3 & 3\\
RRF-Net \cite{liu2017learning} & fixed ResNet-152 & w2v + HGLMM &47.6 & 77.4 & 87.1 & - & 35.4 & 68.3 & 79.9 & - \\
2WayNet \cite{eisenschtat2016linking} &  fixed VGG-16 & feature from \cite{klein2015associating} &49.8 & 67.5 & - & - & 36.0 & 55.6 & - & - \\
DAN (VGG-19) \cite{nam2016dual}  & fixed VGG-19& ft RNN & 41.4 & 73.5 & 82.5 & 2 & 31.8 & 61.7 & 72.5 & 3 \\
DAN (ResNet-152) \cite{nam2016dual}  & fixed ResNet-152& ft RNN & 55.0 & 81.8 & 89.0 & 1 & \textbf{39.4} & 69.2 & 79.1 & 2 \\
\hline
Ours (VGG-19) Stage I  & fixed VGG-19& ft ResNet-50$^\dagger$ (w2v init.)& 37.5 & 66.0 & 75.6 & 3 & 27.2 & 55.4 &  67.6 & 4\\
Ours (VGG-19) Stage II  & ft VGG-19& ft ResNet-50$^\dagger$ (w2v init.)& 47.6 & 77.3 & 87.1 & 2 & 35.3 & 66.6 & 78.2 & 3 \\
Ours (ResNet-50) Stage I  & fixed ResNet-50& ft ResNet-50$^\dagger$ (w2v init.)& 41.2 & 69.7 & 78.9 & 2 & 28.6 & 56.2 & 67.8 & 4\\
Ours (ResNet-50) Stage II  & ft ResNet-50 & ft ResNet-50$^\dagger$ (w2v init.)& 53.9 & 80.9 & \textbf{89.9} & \textbf{1} & 39.2 & \textbf{69.8} & 80.8 & \textbf{2}\\
Ours (ResNet-152) Stage I  & fixed ResNet-152& ft ResNet-152$^\dagger$ (w2v init.)& 44.2 & 70.2 & 79.7 & 2 & 30.7 & 59.2 & 70.8 & 4\\
Ours (ResNet-152) Stage II  & ft ResNet-152& ft ResNet-152$^\dagger$ (w2v init.)& \textbf{55.6} & \textbf{81.9} & 89.5 & \textbf{1} & 39.1 & 69.2 &  \textbf{80.9} & \textbf{2} \\
\hline
\end{tabular}
\end{center}
\caption{Method comparisons on Flickr30k. ``Image Query'' denotes using an image  as query to search for the relavant sentences, and `Text Query' denotes using a sentence to find the relevant image. R@K is Recall@K (higher is better). Med $r$ is the median rank (lower is better). ``ft'' means fine-tuning. $^\dagger$: Text CNN structure is similar to the image CNN, illustrated in Fig. \ref{fig:3}. }
\label{table:Flickr30k}
\end{table*}
\section{Experiment} \label{sec:experiments}
We first introduce the three large-scale image-text retrieval datasets, \ie, Flickr30k, MSCOCO and CUHK-PEDES, followed by the evaluation metric in Section \ref{sec:dataset}. Then Section \ref{sec:details} describes the implementation details and the reproducibility. We discuss the comparison with state of the art and mechanism study in Section \ref{sec:state} and Section \ref{sec:study}.

\subsection{Datasets} \label{sec:dataset}
\textbf{Flickr30k} \cite{young2014image} is one of the large-scale image captioning datasets. It contains 31,783 images collected from Flickr, in which every image is annotated with five text descriptions. The average sentence length is $10.5$ words after removing rare words. We follow the protocol in \cite{hodosh2013framing,karpathy2014deep} to split the dataset into 1,000 test images, 1,000 validation images, and 29,783 training images.

\textbf{MSCOCO} \cite{lin2014microsoft} contains 123,287 images and 616,767 descriptions. Every images contains roughly 5 text descriptions on average. The average length of captions is $8.7$ after rare word removal.  
Following the protocol in \cite{karpathy2015deep}, we randomly select 5,000 images 
as test data and 5,000 images as validation data. The remaining 113,287 images are used as training data. The evaluation is reported on 1K test images (5 fold) and 5K test images. 


\textbf{CUHK-PEDES} \cite{li2017person} collects images from many different person re-identification datasets. It contains 40,206 images from 13,003 different pedestrians and 80,440 descriptions. On average, each person has 3.1 images, and each image has 2 sentences. The average sentence length is $19.6$ words after we remove  rare words.  
We follow the protocol in \cite{li2017person}, selecting the last 1,000 persons for evaluation. There are 3,074 test images with 6,156 captions, 3,078 validation images with 6,158 captions, and 34,054 training images with 68,126 captions. 

\textbf{Evaluation Metric} We use two evaluation metrics \ie, Recall@K and Median Rank. \textbf{Recall@K} is the possibility that the true match appears in the top K of the rank list, where a higher score is better. \textbf{Median Rank} is the median rank of the closest ground truth result in the rank list, with a lower index being better. 

\subsection{Implementation Details} \label{sec:details}
The model is trained by stochastic gradient descent (SGD) with momentum fixed to 0.9 for weight update. While training, the images are resized to $224 \times 224$ pixels which are randomly cropped from images whose shorter size is $256$. 
We also perform simple data augmentation such as horizontal flipping.  For training text input, we conduct position shift (Section \ref{sec:text_cnn}) as data augmentation. Dropout is applied to both CNNs, and the dropout rate is $0.75$. For Flickr30k and MSCOCO, we set the max text length to $32$; for CUHK-PEDES, we set the max text length to $56$, since most sentences are longer. 

In the first training stage, we fixed the pre-trained image CNN, and train the text CNN only. The learning rate is 0.001. We stop training when instance loss converges. In the second stage, we combine the ranking loss as Eq. \ref{eq:final_loss} (the margin $\alpha = 1$) and fine-tune the entire network. 

When testing, we can use the trained image CNN and trained text CNN separately. We extract the image feature $f_{img}$ by image CNN and the text feature $f_{text}$ by text CNN. We use the cosine distance to evaluate the similarity between the query and candidate images/sentences. It is consistent with the similarity used in the ranking loss objective. The final retrieval result is based on the similarity ranking. We also conduct the horizontal flipping when testing and use the average features (no flip and flip) as the image feature. 

\textbf{Reproducibility.} Our source code is available online\footnote{\url{https://github.com/layumi/Image-Text-Embedding}}. The implementation is based on the Matconvnet package \cite{vedaldi15matconvnet}. Since the entire network only uses four components \ie, convolution, pooling, ReLU and batch normalization, it can be easily modified to other deep learning packages. 

\textbf{Training Time} The image CNN (ResNet-50) in our method uses $\sim$119 ms per image batch (batch size = 32) on an Nvidia 1080Ti GPU. The text CNN (similar ResNet-50) also uses $\sim$117 ms per sentence batch (batch size = 32). Therefore, the image feature and text feature can be simultaneously calculated. Although our implementation is sequential, the model can run in a parallel style efficiently.

\subsection{Comparison with State of the Art} \label{sec:state}
We first compare our method with the state-of-the-art methods on the three datasets, \ie, Flickr30k, MSCOCO, and CUHK-PEDES. The compared methods include recent models on the bidirectional image and sentence retrieval. 
For a fair comparison, we present the results based on different image CNN structures, \ie, VGGNet \cite{simonyan2014very} and ResNet \cite{he2016deep}. We also summarise the visual and textual embeddings used in these works in Table \ref{table:Flickr30k} and Table \ref{table:MSCOCO}.  
Extensive results are shown in Table \ref{table:Flickr30k}, Table \ref{table:MSCOCO}, and Table \ref{table:CUHK}, respectively. 
On \textbf{Flickr30k}, we achieve competitive results with state-of-the-art DAN \cite{nam2016dual}: Recall@1 = 55.6\%, Med $r$ = 1 using image queries, and Recall@1 = 39.1\%, Med $r$ = 2 using text queries. While both based on VGG-19, our method exceeds DAN \textbf{$6.2\%$} and \textbf{$3.5\%$} Recall@1 using image and text query respectively.
On \textbf{MSCOCO} 1K-test-image setting, we arrive at Recall@1 = 65.6\%,~Med $r$ = 1 using image queries, and Recall@1 = 47.1\%,~Med $r$ = 2 using text queries. On 5K-test-image setting, we arrive at Recall@1 = 41.2\%, Med $r$ = 2 using image queries, and Recall@1 = 25.3\%, Med $r$ = 5 using text queries.
CUHK-PEDES is a specific dataset for retrieving pedestrian images using the textual description. On \textbf{CUHK-PEDES}, we arrive at Recall@1 = 32.15\%, Med $r$ = 4. While both are based on a VGG-16 network, our model has \textbf{6.21\%} higher recall rate. Moreover, our model based on ResNet-50 achieves new state-of-the-art performance: Recall@1 = 44.4\%, Med $r$ = 2 using language description to search relevant pedestrians. Our method exceeds the second best method \cite{li2017identity} by \textbf{18.46\%} in Recall@1 accuracy. 

Note that m-CNN \cite{ma2015multimodal} also fine-tunes the CNN model to extract visual and textual features. m-CNN encompasses four different levels of text matching CNN while we only use one deep textual model with residual blocks. While both are based on VGG-19, our model has higher performance than m-CNN. Compared with a recent arXiv work, VSE++ \cite{faghri2017vse++},  our result is also competitive.




\setlength{\tabcolsep}{6pt}
\begin{table}
\footnotesize
\begin{center}
\begin{tabular}{l|c|cc|cc}
\hline
\multirow{2}{*}{Method} & \multirow{2}{*}{Stage} & \multicolumn{2}{c|}{Image Query} & \multicolumn{2}{c}{Text Query}\\
  &  & R@1 & R@10 & R@1 & R@10\\
\shline
Only Ranking Loss & I & 6.1 & 27.3 & 4.9 & 27.8\\
Only Instance Loss &  I & 39.9 & 79.1 & 28.2 & 67.9 \\
\hline
Only Instance Loss &  II & 50.5 & 86.0 & 34.9 & 75.7 \\
Only Ranking Loss & II & 47.5 & 85.4 & 29.0 & 68.7 \\
Full model & II & 55.4 & 89.3 & 39.7 & 80.8 \\
\hline
\end{tabular}
\end{center}
\caption{Ranking loss and instance loss retrieval results on Flickr30k validation set. Except for the different losses, we apply the entirely same network (ResNet-50). For a clear comparison, we also fixed the image CNN in Stage I and tune the entire network in Stage II to observe the overfitting. }
\label{table:instance>Rank}
\end{table}


\subsection{Mechanism Study} \label{sec:study}
\textbf{The effect of Stage 1 training.} \label{stage1}
We replace the instance loss with the ranking loss at the first stage when fixing the image CNN. As shown in Table \ref{table:instance>Rank}, the performance is limited. As discussed in Section \ref{discussion}, ranking loss focuses on inter-modal distance. It may be hard to tune the visual and textual features simultaneously at the beginning. As we expected, instance loss performs better, which focuses more on learning intra-modal discriminative descriptors.

\textbf{Two losses can works together.} \label{stage2}
In Stage II, the experiment on the validation set verifies that two losses can work together to improve the final retrieval result (see Table \ref{table:instance>Rank}). Compared with models using only ranking loss or instance loss, the model with two losses provides for higher performance. In the second stage, instance loss does help to regularize the model.

\textbf{End-to-end fine-tuning helps.}
In Stage II, we fine-tune the entire network. For the two general object datasets Flickr30k and MSCOCO, fine-tuning the whole network can improve the rank-1 accuracy by approximately 10\% (see Table. \ref{table:Flickr30k} and Table. \ref{table:MSCOCO}). Imagenet collects images from the Internet, while the pedestrian dataset CUHK-PEDES collects images from surveillance cameras. The fine-tuning result is more obvious on the CUHK-PEDES due to the different data distribution. 
The fine-tuned network (based on ResNet-50) improves the Recall@1 by 29.37\%. The experiments indicate the end-to-end training is critical to image-sentence retrieval, especially person search.

\textbf{Do we really need so many classes?}
For instance loss, the number of classes is usually large. Is it possible to use fewer classes? We implement the pseudo-category method by k-means clustering on MSCOCO, since MSCOCO has most images (classes). We use pool5 feature of ResNet50 pretrained on ImageNet to cluster $3,000$ and $10,000$ categories by K-means. The clustering results are used as the pseudo label for the images to conduct classification. Although clustering can decrease the number of training classes and add the samples per classes, different instances are forced to be of the same class and details may be lost (black / gray dog, two dogs), which compromises the accuracy. The retrieval result with k-classes on MSCOCO is shown in Table \ref{table:R1}. It shows that the strategy is inferior to the instance loss. 

\textbf{Deeper Text CNN does not improve the performance}
Several previous works report the Text CNN may not improve the result when the network is very deep \cite{le2017convolutional,conneau2017very}. It is different with the observation in the image recognition \cite{he2016deep}. In our experiment, we also observe a similar result when deepening the Text CNN on Flickr30k and MSCOCO. Deeper Text CNN does not significantly improve the result (see Table \ref{table:deeper}). 

\setlength{\tabcolsep}{3pt}
\begin{table}
\footnotesize
\begin{center}
\begin{tabular}{l|c|c|c}
\hline
Methods  & Dataset &Image-Query R@1 &Text-Query R@1\\
\shline
Res152 + Res50$^\dagger$  & \multirow{2}{*}{Flickr30k} & 44.4 & 29.6 \\
Res152 + Res152$^\dagger$  & & 44.2 & 30.7 \\
\hline
Res152 + Res50$^\dagger$  & \multirow{2}{*}{MSCOCO}& 52.0& 38.0\\
Res152 + Res152$^\dagger$  & & 52.8 & 37.7\\
\hline
\end{tabular}
\end{center}
\caption{Deeper Text CNN on Flickr30k and MSCOCO. We use fixed Image CNN (StageI). $^\dagger$: Text CNN structure.
}
\label{table:deeper}
\end{table}

\setlength{\tabcolsep}{7pt}
\begin{table}
\footnotesize
\begin{center}
\begin{tabular}{l|c|c}
\hline
Methods  &Image-Query R@1 &Text-Query R@1\\
\shline
3000 categories (StageI)  & 38.0 & 26.1 \\
10000 categories (StageI)  & 44.7 & 31.3 \\
Our (StageI)  & 52.2 & 37.2 \\
\hline
\end{tabular}
\end{center}
\caption{K-class Loss vs. Instance Loss on MSCOCO. We use the K-means clustering result as pseudo categories. The experiment is based on Res50 + Res50$^\dagger$ as the model structure.}
\label{table:R1}
\end{table}

\setlength{\tabcolsep}{7pt}
\begin{table}
\footnotesize
\begin{center}
\begin{tabular}{l|cc|cc}
\hline
\multirow{2}{*}{Method} & \multicolumn{2}{c|}{Image Query} & \multicolumn{2}{c}{Text Query}\\
  & R@1 & R@10 & R@1 & R@10\\
\shline
Random initialization \cite{glorot2010understanding} & 38.0 & 78.7 & 26.6 & 66.6 \\
Word2vec initialization & 39.9 & 79.1 & 28.2 & 67.9 \\
\hline
\end{tabular}
\end{center}
\caption{Ablation study. With/without word2vec initialization on Flickr30k validation. 
The result suggests \emph{word2vec} serves as a proper initialization for text CNN.
}
\label{table:Word2vec}
\end{table}

\setlength{\tabcolsep}{10pt}
\begin{table}
\footnotesize
\begin{center}
\begin{tabular}{l|cc|cc}
\hline
\multirow{2}{*}{Method} & \multicolumn{2}{c|}{Image Query} & \multicolumn{2}{c}{Text Query}\\
  & R@1 & R@10 & R@1 & R@10\\
\shline
Left alignment & 34.1 & 73.1 & 23.6 & 61.4 \\
Position shift & 39.9 & 79.1 & 28.2 & 67.9 \\
\hline
\end{tabular}
\end{center}
\caption{Ablation study. Position shift vs. Left alignment on Flickr30k validation. 
It shows that position shift can serve as a significant data augmentation method for the text CNN.
}
\label{table:Position shift}
\end{table}

\setlength{\tabcolsep}{4pt}
\begin{table}
\footnotesize
\begin{center}
\begin{tabular}{l|c|cccc}
\hline
\multirow{2}{*}{Method} & \multirow{2}{*}{Visual} &\multicolumn{4}{c}{Text Query}\\
 & & R@1 & R@5 & R@10 & Med $r$\\
\shline
CNN-RNN (VGG-16$^\ddagger$) \cite{reed2016learning} & fixed & 8.07 & - & 32.47 & -\\
Neural Talk (VGG-16$^\ddagger$) \cite{vinyals2015show} & fixed & 13.66 & - & 41.72 & -\\
GNA-RNN (VGG-16$^\ddagger$) \cite{li2017person} & fixed & 19.05 & - & 53.64 & -\\
IATV (VGG-16) \cite{li2017identity} & ft & 25.94 & - & 60.48 & - \\
\hline
Ours (VGG-16) Stage I & fixed & 14.26 & 33.07 & 43.47 & 16 \\
Ours (VGG-16) Stage II & ft & 32.15 & 54.42 & 64.30 & 4 \\
Ours (ResNet-50) Stage I & fixed & 15.03 & 31.66 & 41.62 & 18  \\
Ours (ResNet-50) Stage II & ft & \textbf{44.40} & \textbf{66.26} & \textbf{75.07} & \textbf{2} \\
\hline
\end{tabular}
\end{center}
\caption{Method comparisons on CUHK-PEDES. R@K (\%) is Recall@K (high is good). Med $r$ is the median rank (low is good). ft means fine-tuning. $^\ddagger$: pre-trained on person identification.}
\label{table:CUHK}
\end{table}

\setlength{\tabcolsep}{5pt}
\begin{table*}
\footnotesize
\begin{center}
\begin{tabular}{l|c|c|cccc|cccc}
\hline
\multirow{2}{*}{Method} & \multirow{2}{*}{Visual} & \multirow{2}{*}{Textual} &\multicolumn{4}{c|}{Image Query} & \multicolumn{4}{c}{Text Query}\\
  & &  & R@1 & R@5 & R@10 & Med & R@1 & R@5 & R@10 & Med $r$\\
\shline
1K test images\\
\shline
DVSA \cite{karpathy2015deep} & ft RCNN & w2v + ft RNN& 38.4 & 69.9 & 80.5 & 1 & 27.4 & 60.2 & 74.8 & 3 \\
GMM-FV \cite{klein2015associating} & fixed VGG-16 & w2v + GMM + HGLMM& 39.4 & 67.9 & 80.9 & 2 & 25.1 & 59.8 & 76.6 & 4\\
m-RNN \cite{mao2014deep}  & fixed VGG-16 & ft RNN & 41.0 & 73.0 & 83.5 & 2 & 29.0 & 42.2 & 77.0 & 3 \\
RNN-FV \cite{lev2016rnn}  & fixed VGG-19 & feature from \cite{klein2015associating} & 41.5 & 72.0 & 82.9 & 2 & 29.2 & 64.7 & 80.4 & 3 \\
m-CNN \cite{ma2015multimodal}  & ft VGG-19 & 4 $\times$ ft CNN& 42.8 & 73.1 & 84.1 & 2 & 32.6 & 68.6 & 82.8 & 3\\
HM-LSTM \cite{niu2017hierarchical}  & fixed CNN from \cite{karpathy2015deep}& ft RNN& 43.9 & - & 87.8 & 2 & 36.1 & - & 86.7 & 3 \\
SPE \cite{wang2016learning}  & fixed VGG-19 & w2v + HGLMM& 50.1 & 79.7 & 89.2 & - & 39.6 & 75.2 & 86.9 & - \\
VQA-A \cite{lin2016leveraging} & fixed VGG-19 & ft RNN & 50.5 & 80.1 & 89.7 & - & 37.0 & 70.9 & 82.9 & - \\
sm-LSTM \cite{huang2016instance} & fixed VGG-19 & ft RNN & 53.2 & 83.1 & 91.5 & 1 & 40.7 & 75.8 & 87.4 & 2\\
2WayNet \cite{eisenschtat2016linking} &  fixed VGG-16 & feature from \cite{klein2015associating} & 55.8 & 75.2 & - & - & 39.7 & 63.3 & - & - \\
RRF-Net \cite{liu2017learning} & fixed ResNet-152 & w2v + HGLMM & 56.4 & 85.3 & 91.5 & - & 43.9 & 78.1 & 88.6 & - \\
\hline
Ours (VGG-19) Stage I & fixed VGG-19 & ft ResNet-50$^\dagger$ (w2v init.)& 46.0 & 75.6 & 85.3 & 2 & 34.4 & 66.6 & 78.7 & 3 \\
Ours (VGG-19) Stage II & ft VGG-19 & ft ResNet-50$^\dagger$ (w2v init.)& 59.4 & 86.2 & 92.9 & 1 & 41.6 & 76.3 & 87.5 & 2 \\
Ours (ResNet-50) Stage I &fixed ResNet-50 & ft ResNet-50$^\dagger$ (w2v init.)& 52.2 & 80.4 & 88.7 & 1 & 37.2 & 69.5 & 80.6 & 2 \\
Ours (ResNet-50) Stage II & ft ResNet-50 & ft ResNet-50$^\dagger$ (w2v init.)& \textbf{65.6} & \textbf{89.8} & \textbf{95.5} & \textbf{1} & \textbf{47.1} & \textbf{79.9} & \textbf{90.0} &  \textbf{2}  \\
\shline
5K test images\\
\shline
GMM-FV \cite{klein2015associating} & fixed VGG-16 & w2v + GMM + HGLMM  & 17.3 & 39.0 & 50.2 & 10 & 10.8 & 28.3 & 40.1 & 17\\
DVSA \cite{karpathy2015deep} & ft RCNN & w2v + ft RNN & 16.5 & 39.2 & 52.0 & 9 & 10.7 & 29.6 & 42.2 & 14 \\
VQA-A \cite{lin2016leveraging} & fixed VGG-19 & ft RNN  & 23.5 & 50.7 & 63.6 & - & 16.7 & 40.5 & 53.8 & - \\
\hline
Ours (VGG-19) Stage I  & fixed VGG-19 & ft ResNet-50$^\dagger$ (w2v init.)& 24.5 & 50.1 & 62.1 & 5 & 16.5 & 39.1 & 51.8 & 10 \\
Ours (VGG-19) Stage II  & ft VGG-19 & ft ResNet-50$^\dagger$ (w2v init.)& 35.5 & 63.2 & 75.6 & 3 & 21.0 & 47.5 & 60.9 & 6 \\
Ours (ResNet-50) Stage I  & fixed ResNet-50 & ft ResNet-50$^\dagger$ (w2v init.)& 28.6 & 56.2 & 68.0 & 4 & 18.7 & 42.4 & 55.1 & 8 \\
Ours (ResNet-50) Stage II  & ft ResNet-50 & ft ResNet-50$^\dagger$ (w2v init.) & \textbf{41.2} & \textbf{70.5} & \textbf{81.1} & \textbf{2} & \textbf{25.3} & \textbf{53.4} & \textbf{66.4} & \textbf{5} \\
\hline
\end{tabular}
\end{center}
\vspace{-.05in}
\caption{Method comparisons on MSCOCO. R@K (\%) is Recall@K (high is good). Med $r$ is the median rank (low is good). 1K test images denotes using five non-overlap splits of 5K images to conduct retrieval evaluation and report the average result. 5K test images means using all images and texts to perform retrieval. ft means fine-tuning. $^\dagger$: Text CNN structure is similar to the image CNN, illustrated in Fig. \ref{fig:3}.}
\vspace{-.1in}
\label{table:MSCOCO}
\end{table*}

\begin{figure}[t]
\begin{center}
   \includegraphics[width=1\linewidth]{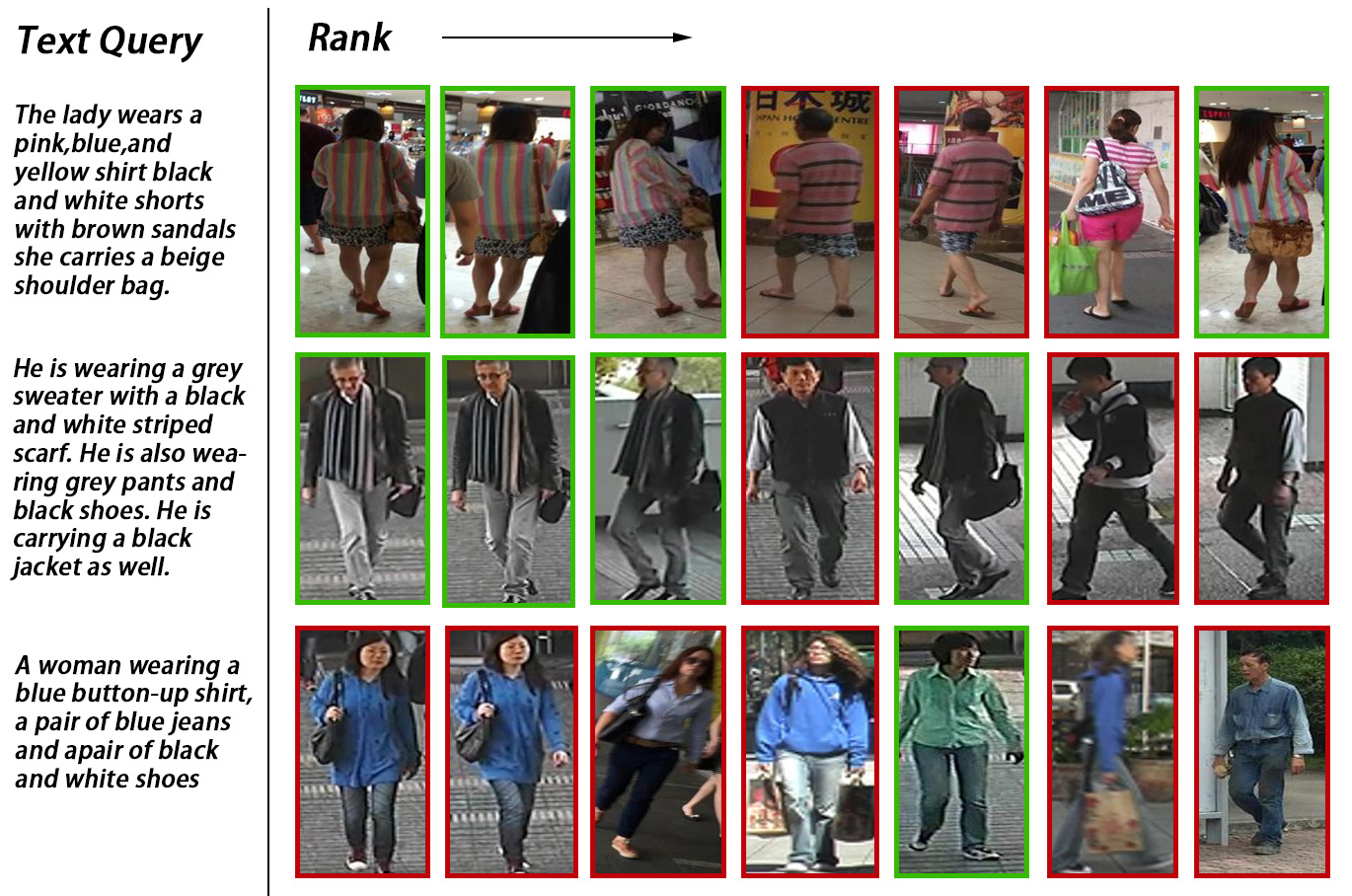}
\end{center}
   \caption{Qualitative image search results using text query. The results are sorted from left to right according to their confidence. The images in green boxes are the true matches, and the images in red boxes are the false matches. In the last row, the rank-1 woman also wears a blue shirt, a pair of blue jeans and a pair of white shoes. The model outputs reasonable false matches. }
\label{fig:CUHK}
\end{figure}

\begin{figure}[t]
\begin{center}
   \includegraphics[width=1\linewidth]{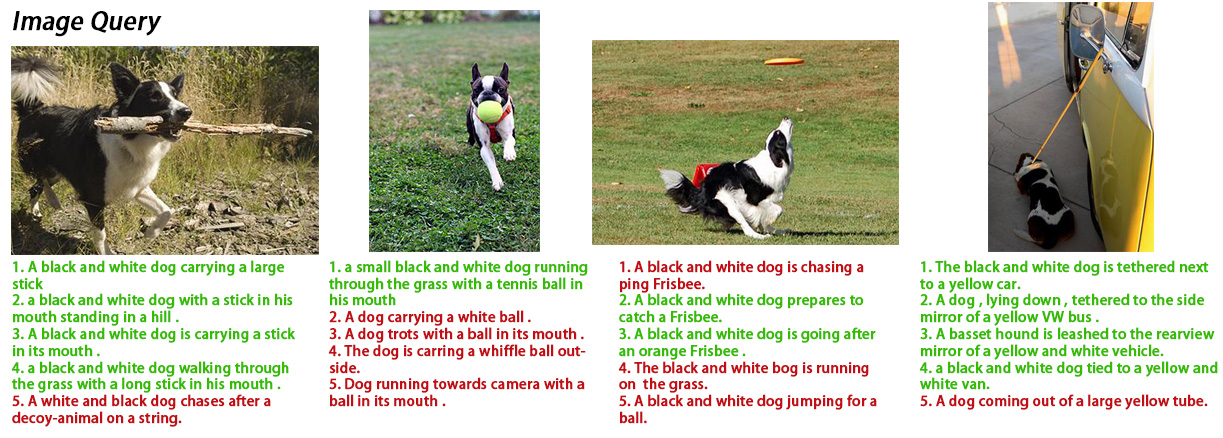}
\end{center}
   \caption{Qualitative description search results using image query on Flickr30k. Below each image we show the top five retrieval sentences (there are 5,000 candidate sentences in the gallery) in descending confidence. Here we select four black and white dogs as our query. Except for the main object (dog), we show the model can correctly recognize environment and small object. The sentences in green are the true matches, and the descriptions in red are the false matches. Note that some general descriptions are also reasonable. (Best viewed when zoomed in.)}
\label{fig:dogs}
\end{figure}

\textbf{Word2vec initialization helps.} 
We compare the result using the \emph{word2vec} initialization or random initialization \cite{glorot2010understanding} for the first convolution layer of text CNN. Note that we remove the words, which have not appeared in the training set, in the training data as well as dictionary. So the weight of first convolution layer is $d \times 300$ instead of $3,000,000 \times 300$. $d$ is the dictionary size. When testing, the missing words in the dictionary will also be removed in advance. As shown in Table. \ref{table:Word2vec}, it can be observed that using \emph{word2vec} initialization outperforms by $1\%$ to $2\%$ compared to the random initialization. Although \emph{word2vec} is not trained on the target dataset, it still serves as a proper initialization for text CNN.

\textbf{Position shift vs. Left alignment}: 
Text CNN has a fixed-length input. As discussed in Section \ref{sec:text_cnn}, left alignment is to pad zeros at the end of text input (like aligning the whole sentence left), if the length of the sentence is shorter than $32$. Position shift is to add zeros at the end of text input as well as the begining of the input. We conduct the position shift online when reading data from the disk. We do the experiment on Flickr30k validation set. As shown in Table \ref{table:Position shift}, the model using position shift outperforms the one using left alignment $\sim5\%$. Position shift serves as a significant data augmentation method for text feature learning. 

In Fig. \ref{fig:CUHK} and Fig. \ref{fig:dogs}, we present some visual retrieval results on CUHK-PEDES  and  Flickr30k, respectively. Our method returns reasonable rank lists. (More qualitative results can be found in Appendix.)


\textbf{Does Text CNN learn discriminative words?} \label{word}
The text CNN is supposed to convey the necessary textual information for image-text matching. To examine whether the text CNN discovers discriminative words, we fix the visual feature. For text input, we remove one word from the sentence each time. If we remove a discriminative word, the matching confidence will drop. In this way, we can determine the learned importance of different words. 

\begin{figure}[t]
\begin{center}
   \includegraphics[width=1\linewidth]{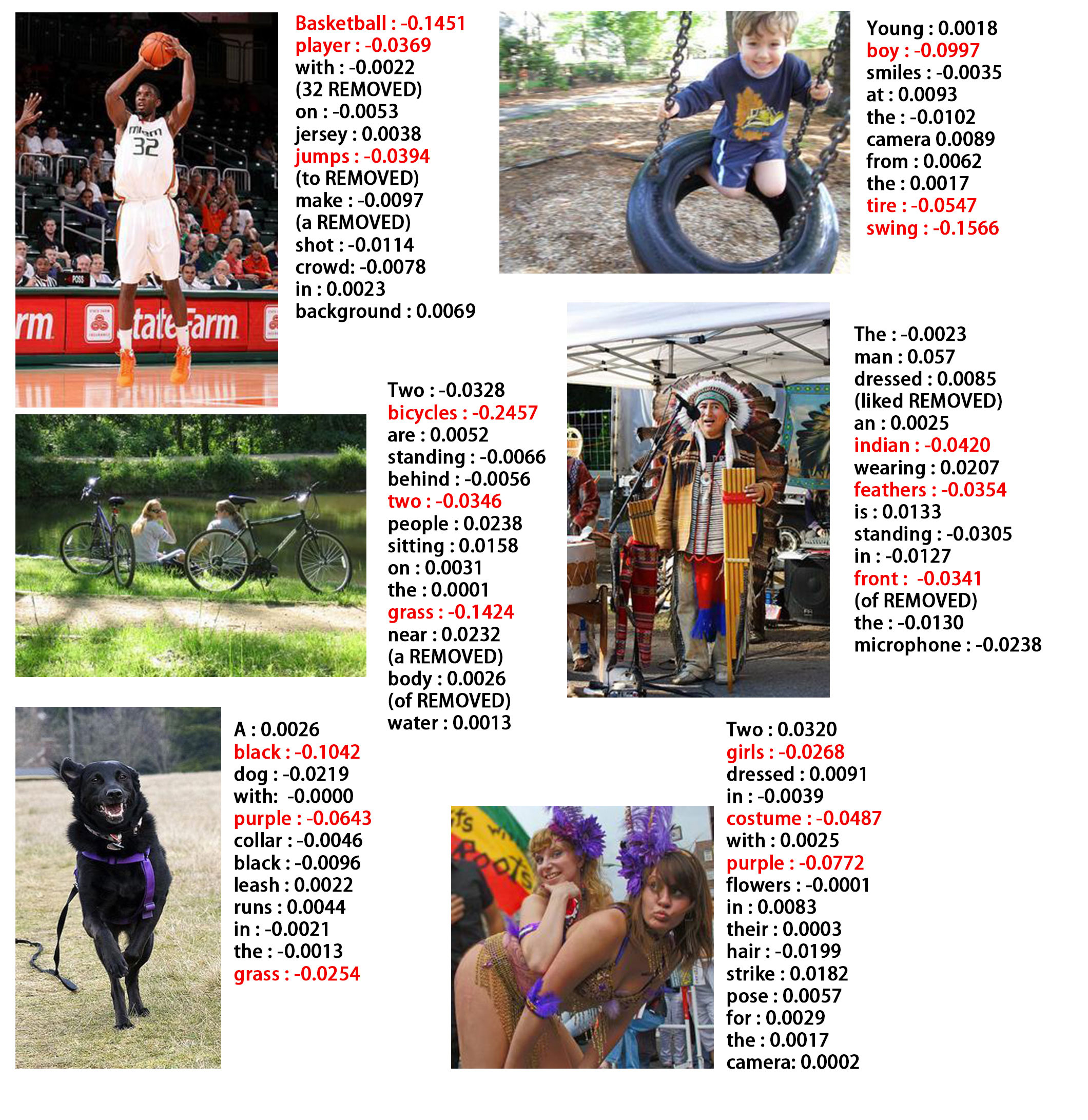}
\end{center}
\vspace{-.1in}
   \caption{Finding discriminative words on Flickr30k test set. Top-3 discriminative words are in red. Some words, which are not in \emph{word2vec} vocabulary, are removed in advance.}
\label{fig:key-word}
\end{figure}

The proposed model learns discriminative words. As show in Fig. \ref{fig:key-word}, we observe that the words which convey the objective/colour information, \ie, \emph{basketball, swing, purple}, are usually discriminative. If we remove these words, the matching confidence drops. Conversely, the conjunctions, \ie, \emph{with, on, at, in}, after being removed, have a small impact on the matching confidence. 

\section{Conclusion} \label{sec:conclusion}
In this paper, we propose the instance loss for image-text retrieval. It is based on an unsupervised assumption that every image/test group can be viewed as one class. The experiment shows instance loss can provide a proper initialization for ranking loss and further regularize the training. 
As a minor contribution, we propose a dual-path CNN to conduct end-to-end training on both image and text branches. The proposed method achieves competitive results on two generic retrieval datasets Flickr30k and MSCOCO. Furthermore, we arrive a +18\% improvement on the person retrieval dataset CUHK-PEDES. Our code has been made publicly available.
Additional examples can be found in Appendix.

\textbf{Acknowledgement.}  Dr Liang Zheng is the recipient of the SIEF STEM+ Bussiness
fellowship


{
\footnotesize
\bibliographystyle{IEEEtran}
\bibliography{egbib}
}

\begin{IEEEbiography}[{\includegraphics[width=1in,height=1.25in,clip,keepaspectratio]{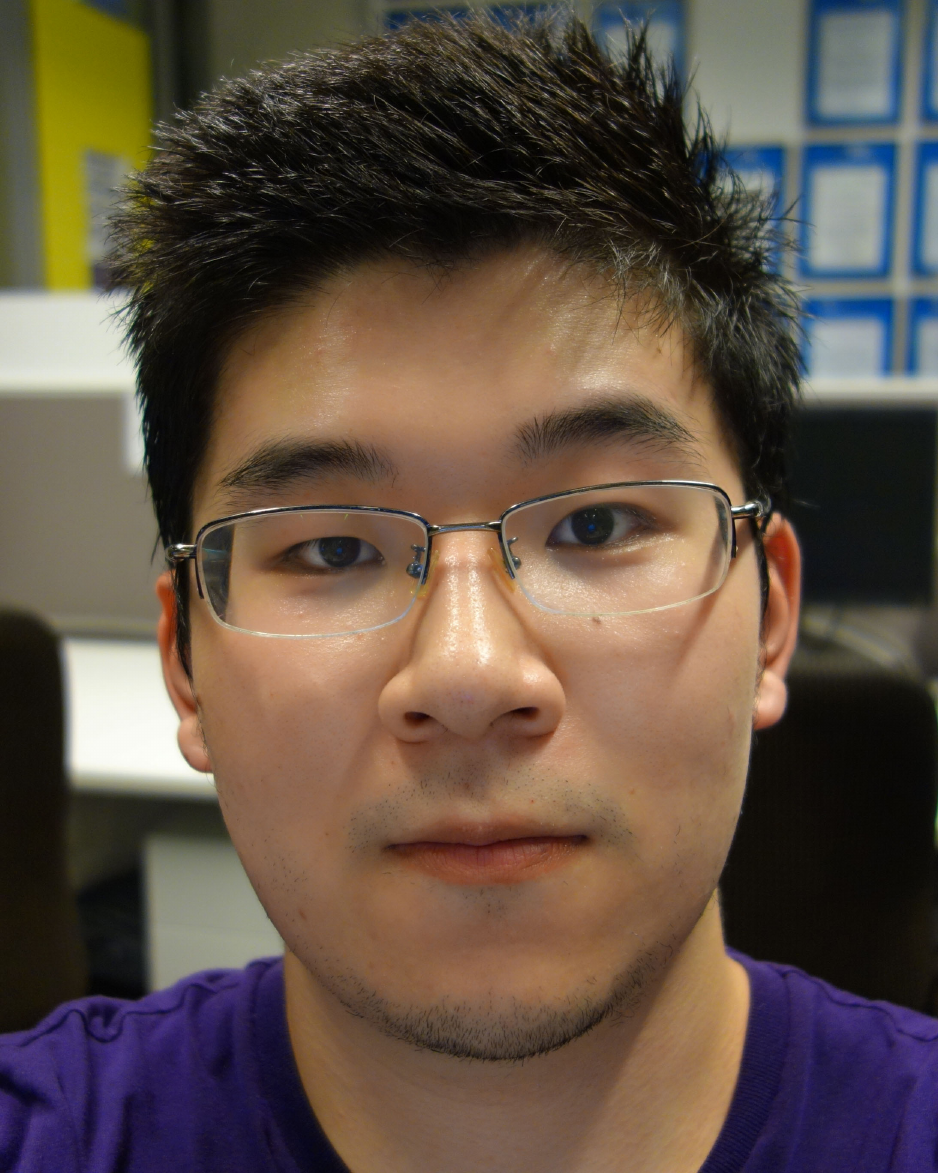}}]{Zhedong Zheng}
received the B.S. degree in Fudan University, China, in 2016. He is currently a Ph.D. student in University of Technology Sydney, Australia. His research interests include image retrieval and person re-identification.
\end{IEEEbiography}
\vfill
\begin{IEEEbiography}[{\includegraphics[width=1in,height=1.25in,clip,keepaspectratio]{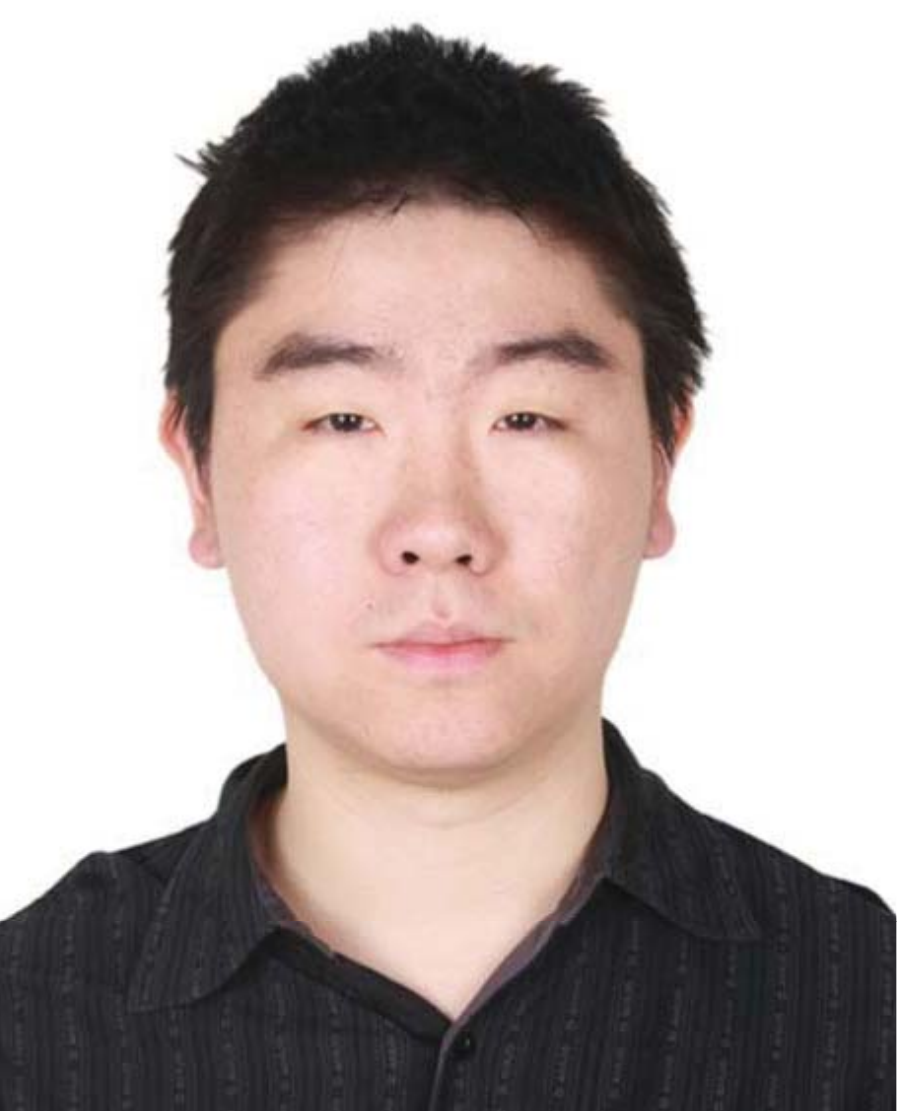}}]{Liang Zheng}
received the Ph.D degree in Electronic Engineering from Tsinghua University, China, in 2015, and the  B.E. degree in Life Science from Tsinghua University, China, in 2010. He was a postdoc researcher in University of Texas at San Antonio, USA. He is currently a postdoc researcher in Quantum Computation and Intelligent Systems, University of Technology Sydney, Australia. His research interests include image retrieval, classification, and person re-identification.
\end{IEEEbiography}
\vfill
\begin{IEEEbiography}[{\includegraphics[width=1in,height=1.25in,clip,keepaspectratio]{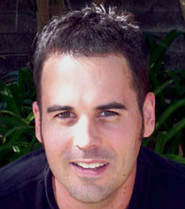}}]{Michael Garrett} received his Ph. D. degree in Educational Technologies from Edith Cowan University, Western Australia in 2012 focusing on the use of three-dimensional graphics applications and simulated learning environments. Michael has undertaken research projects in collaboration with the Australian mining sector and the Royal Australian Navy and has industry experience in software and systems development roles as well as lecturing, research, and consultancy positions within the tertiary education sector. Currently, Michael is the Head of Research \& Innovation at CingleVue International where areas of research interest include educational data mining, machine learning, data analytics and visualisation, virtual and augmented reality technologies, and adaptive and differentiated learning systems.
\end{IEEEbiography}
\vfill
\begin{IEEEbiography}[{\includegraphics[width=1in,height=1.25in,clip,keepaspectratio]{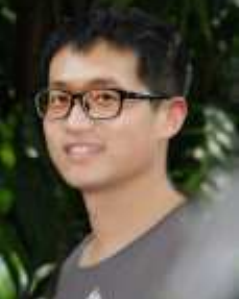}}]{Yi Yang} received the Ph.D. degree in computer
science from Zhejiang University, Hangzhou, China, in 2010. He is currently an associate professor with University of Technology Sydney, Australia.
He was a Post-Doctoral Research with the School of Computer Science, Carnegie Mellon University, Pittsburgh, PA, USA. His current research interest includes machine learning and its applications to multimedia content analysis and computer vision, such as multimedia indexing and retrieval, surveillance video analysis and video semantics understanding.
\end{IEEEbiography}
\vfill
\begin{IEEEbiography}[{\includegraphics[width=1in,height=1.25in,clip,keepaspectratio]{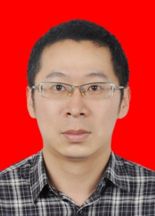}}]{Mingliang Xu}  is a professor in the School of Information Engineering of Zhengzhou University, China. He received his Ph.D. degree in computer science and technology from the State Key Lab of CAD\&CG at Zhejiang University, Hangzhou, China, and the B.S. and M.S. degrees from the Computer Science Department, Zhengzhou University, Zhengzhou, China, respectively. 
\end{IEEEbiography}
\vfill
\begin{IEEEbiography}[{\includegraphics[width=1in,height=1.25in,clip,keepaspectratio]{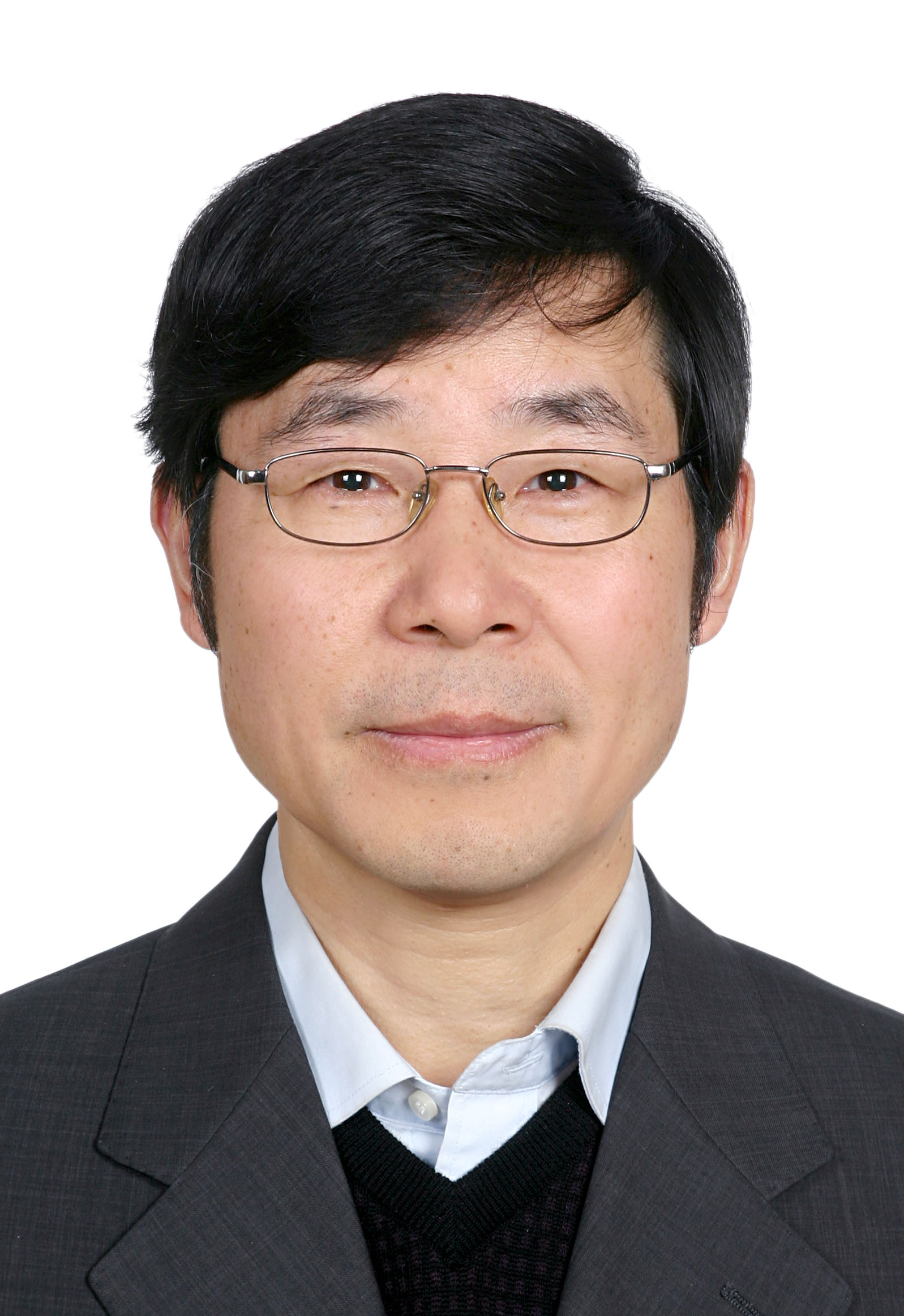}}]{Yi-dong Shen} is a professor of computer science in the State Key Laboratory of Computer Science at the Institute of Software, the Chinese Academy of Sciences, China. Prior to joining this laboratory, he was a professor at Chongqing University, China. His main research interests include Artificial Intelligence and Data Mining.
\end{IEEEbiography}

\begin{figure*}[h]
\begin{center}
   \includegraphics[width=0.8\linewidth]{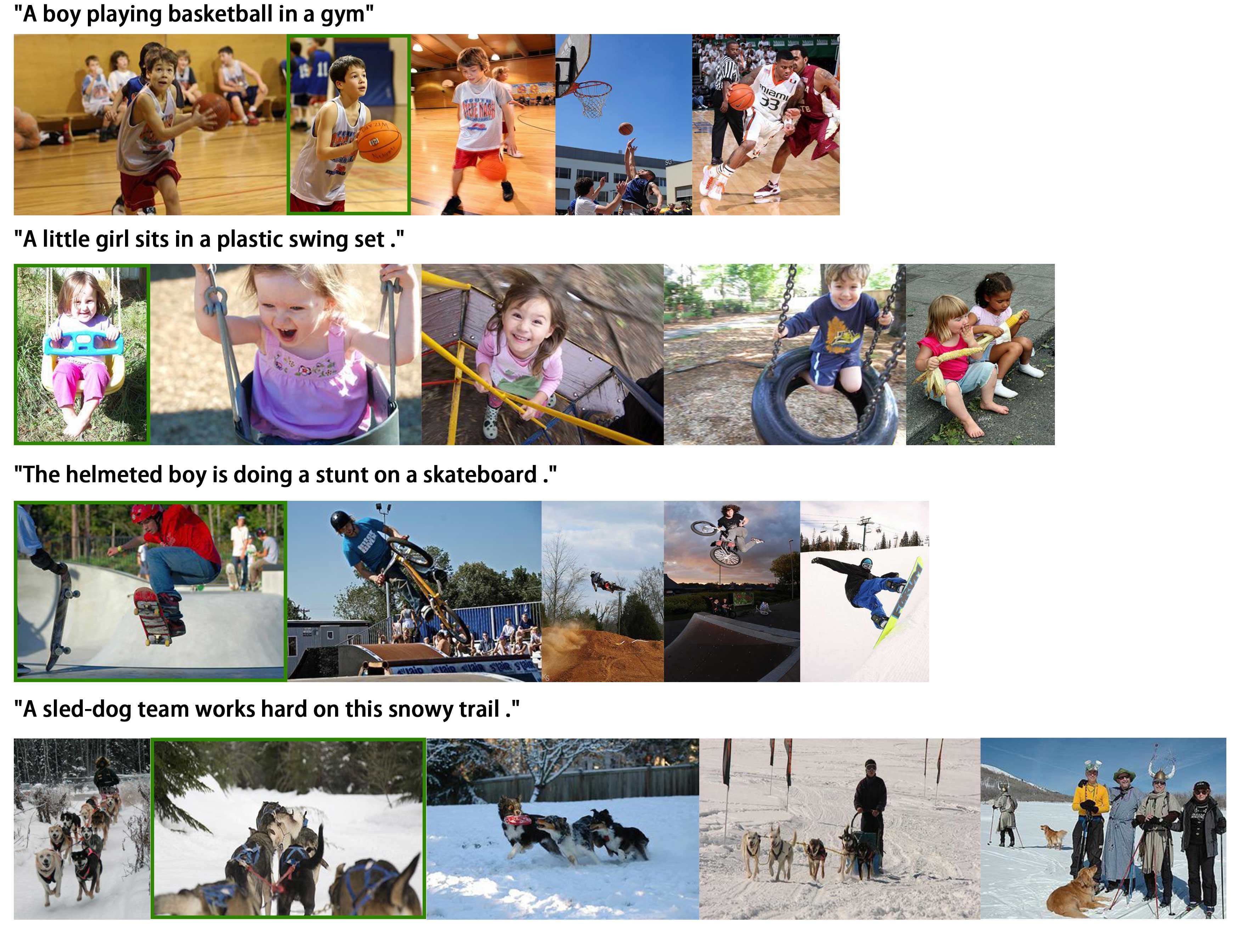}
\end{center}
\vspace{-.2in}
   \caption{ Additional examples of image search (using text queries) on Flickr30k. Top-5 results are sorted from left to right according to their confidence. True matches are in green.}
\vspace{-.2in}
\label{fig:Flickr30k-text2image}
\end{figure*}

\begin{figure*}[h]
\begin{center}
   \includegraphics[width=0.8\linewidth]{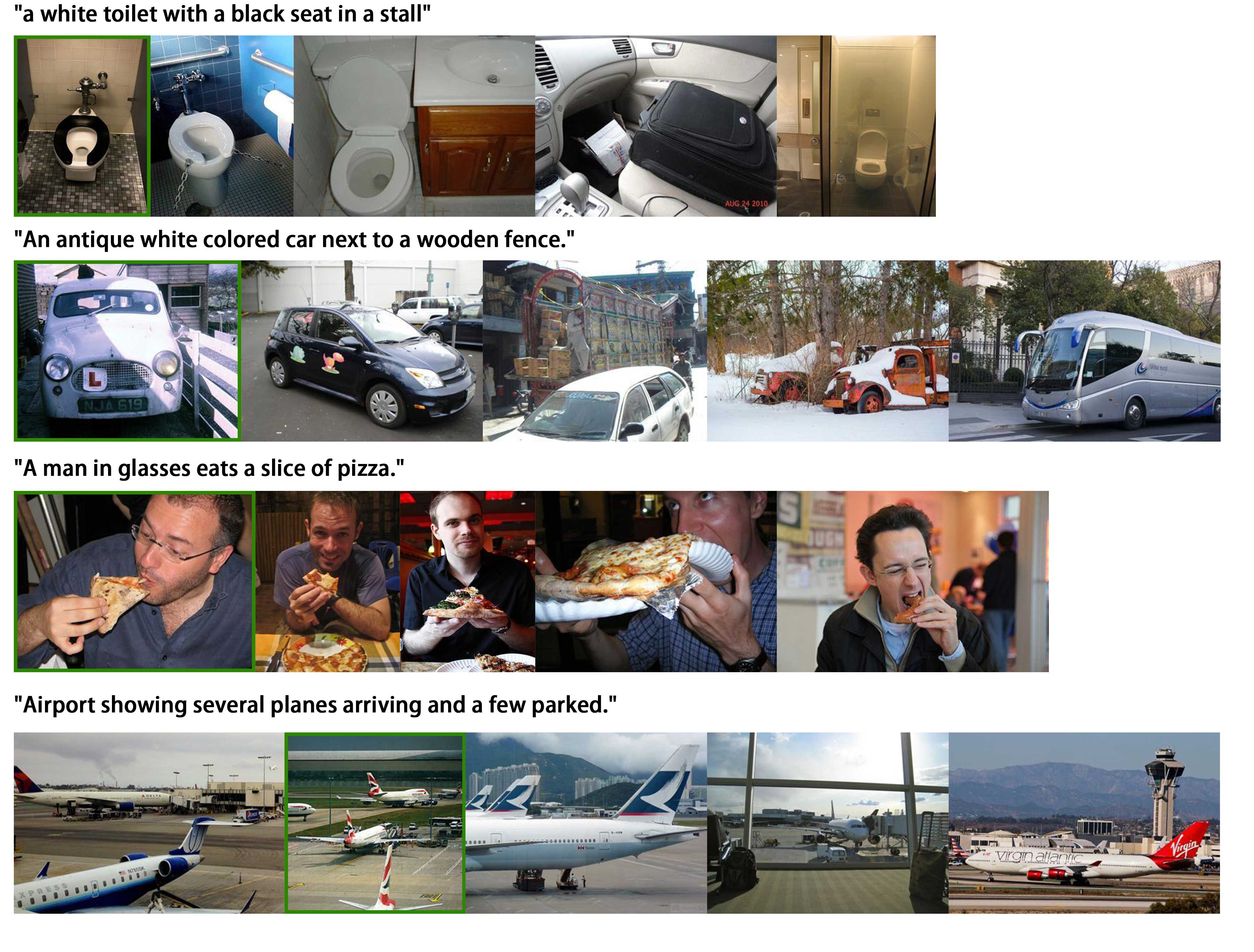}
\end{center}
\vspace{-.2in}
   \caption{ Additional examples of image search (using text queries) on MSCOCO. Top-5 results are sorted from left to right according to their confidence. True matches are in green.}
\vspace{-.2in}
\label{fig:MSCOCO-text2image}
\end{figure*}

\begin{figure*}[h]
\begin{center}
   \includegraphics[width=1\linewidth]{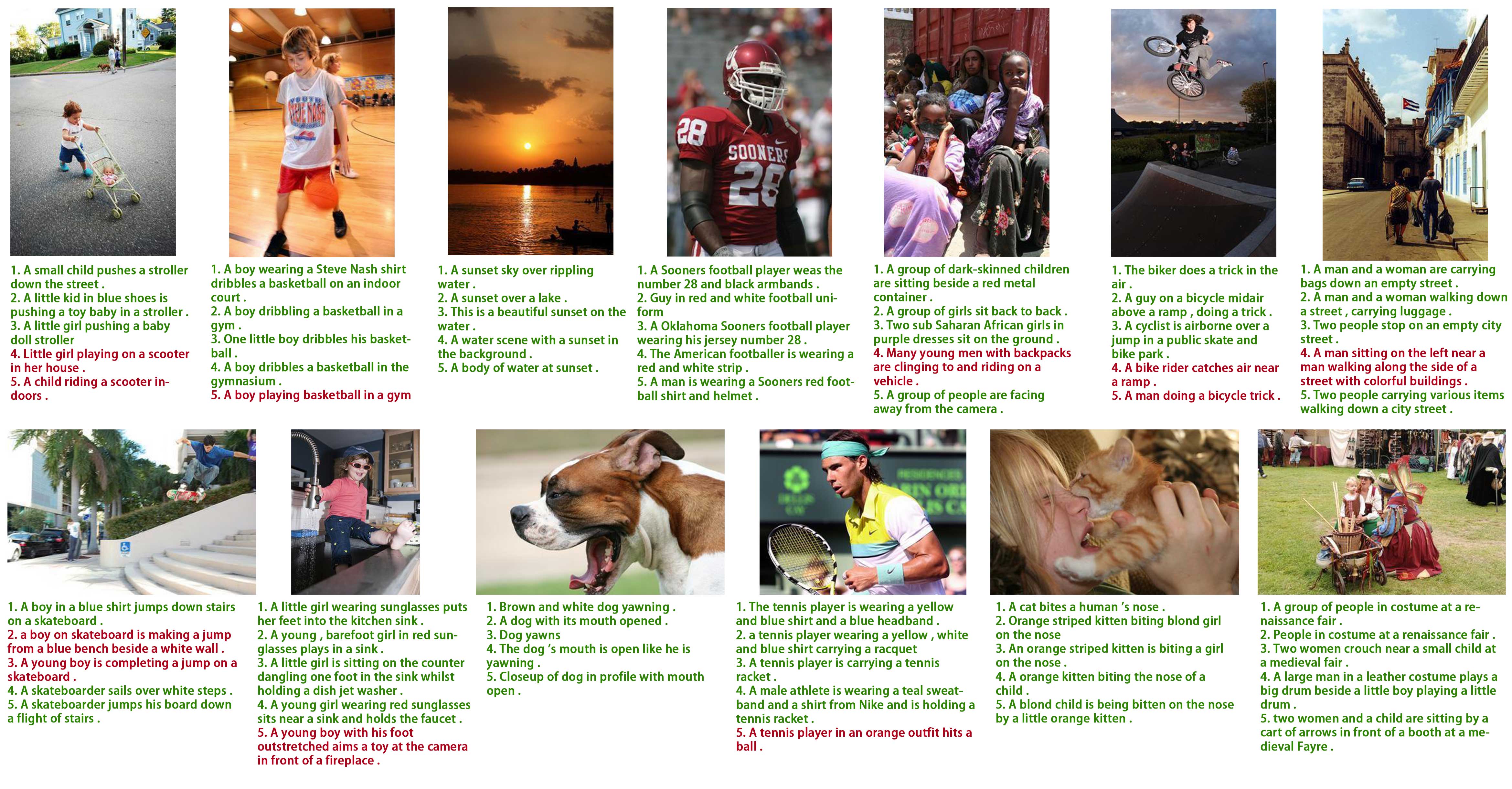}
\end{center}
\vspace{-.1in}
   \caption{ Additional examples of text search (using image queries) on Flickr30k. Under each query image, we show the top five retrieved sentences in descending confidence. The descriptions in green are true matches, and the sentences in red are false matches.}
\vspace{-.1in}
\label{fig:Flickr30k-image2text}
\end{figure*}

\begin{figure*}[h]
\begin{center}
   \includegraphics[width=1\linewidth]{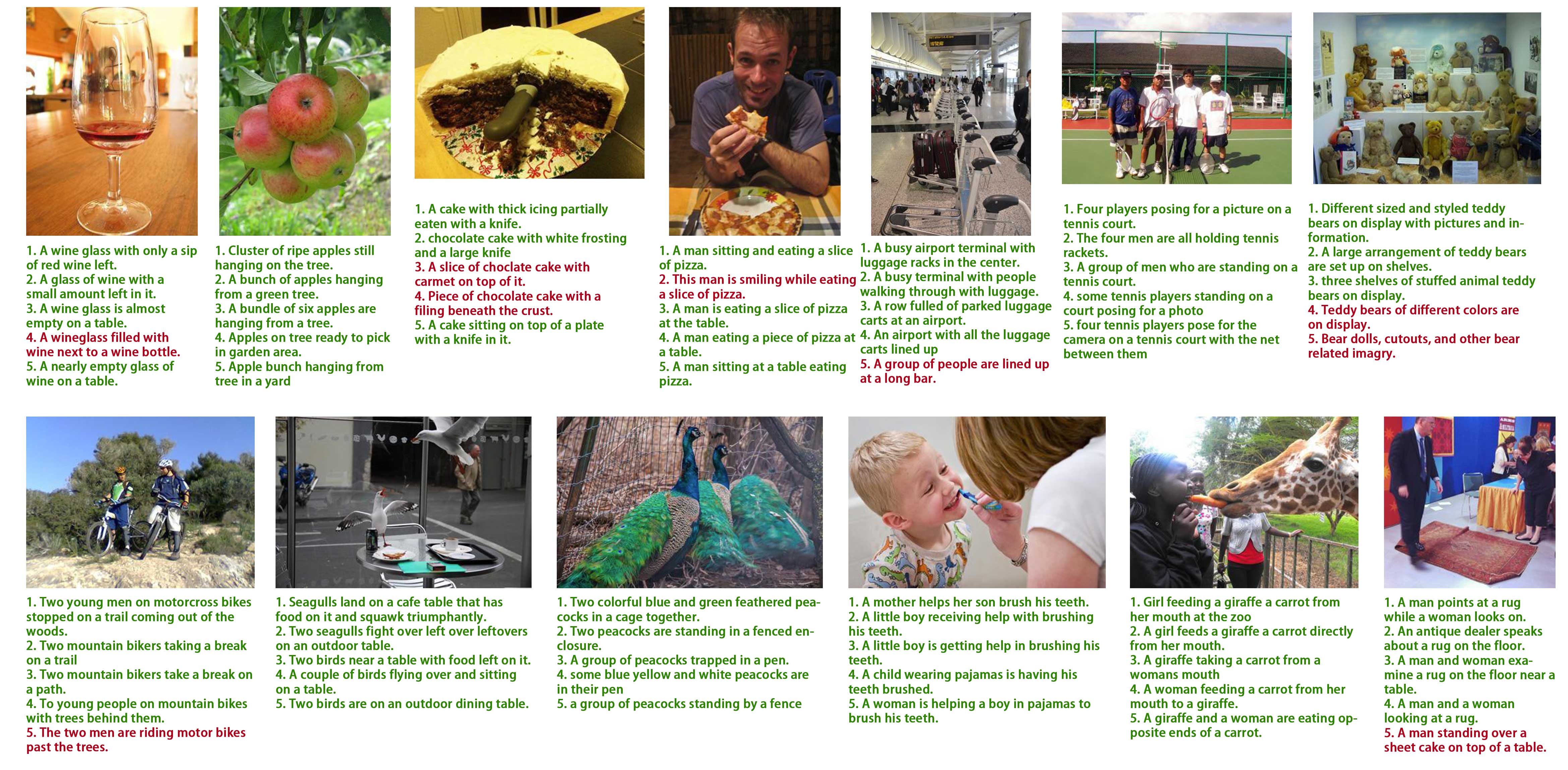}
\end{center}
\vspace{-.1in}
   \caption{ Additional examples of text search (using image queries) on MSCOCO. Under each image, we show the top five retrieval sentences in descending confidence. The descriptions in green are true matches, and the sentences in red are false matches.}
\vspace{-.1in}
\label{fig:MSCOCO-image2text}
\end{figure*}

\appendix
Additional examples of image-text bidirectional retrieval can be found in Fig. \ref{fig:Flickr30k-text2image}, Fig. \ref{fig:MSCOCO-text2image}, Fig. \ref{fig:Flickr30k-image2text} and Fig.\ref{fig:MSCOCO-image2text}. The true matches are in green. 

\end{document}